%% file: main.tex
\documentclass[paper=a4, abstract=true, twocolumn=true, twoside=true, fontsize=10pt]{scrartcl}

\input{preamble}
\input{notation}

\cohead{A Comparative Study of Graph Matching Algorithms in Computer Vision}
\cehead{
  S.~Haller,
  L.~Feineis,
  L.~Hutschenreiter,
  F.~Bernard,
  C.~Rother,
  D.~Kainmüller,
  P.~Swoboda,
  B.~Savchynskyy
}
\ofoot*{}
\ifoot*{}

\title{A Comparative Study of Graph Matching Algorithms in Computer Vision}

\author{
  Stefan Haller\footnotemark[1],\hspace{1em}%
  Lorenz Feineis\footnotemark[1],\hspace{1em}%
  Lisa Hutschenreiter\footnotemark[1],\hspace{1em}%
  Florian Bernard\footnotemark[2],\\%
  Carsten Rother\footnotemark[1],\hspace{1em}%
  Dagmar Kainmüller\footnotemark[3],\hspace{1em}%
  Paul Swoboda\footnotemark[4],\hspace{1em}%
  Bogdan Savchynskyy\footnotemark[1]\\[.8ex]%
  \large
  \footnotemark[1]\, Heidelberg University,
  \footnotemark[2]\, University of Bonn,
  \footnotemark[3]\, MDC Berlin,
  \footnotemark[4]\, MPI Informatik Saarbrücken
}

\date{}

\begin{document}

\maketitle

\begin{abstract}
  \setlength\parindent{15pt}%
  The graph matching optimization problem is an essential component for many tasks in computer vision, such as bringing two deformable objects in correspondence.
  Naturally, a wide range of applicable algorithms have been proposed in the last decades.
  Since a common standard benchmark has not been developed, their performance claims are often hard to verify as evaluation on differing problem instances and criteria make the results incomparable.
  To address these shortcomings, we present a comparative study of graph matching algorithms.
  We create a uniform benchmark where we collect and categorize a large set of existing and publicly available computer vision graph matching problems in a common format.
  At the same time we collect and categorize the most popular open-source implementations of graph matching algorithms.
  Their performance is evaluated in a way that is in line with the best practices for comparing optimization algorithms.
  The study is designed to be reproducible and extensible to serve as a valuable resource in the future.

  Our study provides three notable insights:
  \begin{Itemize}
  \Item popular problem instances are exactly solvable in substantially less than 1 second, and, therefore, are insufficient for future empirical evaluations;
  \Item the most popular baseline methods are highly inferior to the best available methods;
  \Item despite the NP-hardness of the problem, instances coming from vision applications are often solvable in a few seconds even for graphs with more than 500 vertices.
  \end{Itemize}
\end{abstract}

\section{Introduction}
\label{sec:intro}

Finding correspondences between elements of two discrete sets, such as keypoints in images or vertices of 3D meshes, is a fundamental problem in computer vision and as such highly relevant for numerous vision tasks, including 3D reconstruction~\cite{ma2021image}, tracking~\cite{yilmaz2006object}, shape model learning~\cite{heimann2009statistical}, and image alignment~\cite{bernard2015solution}, among others.
Graph matching~\cite{foggia2014graph,sun_survey_2020,yan2016short} is a standard way to address such problems.
In graph matching, vertices of the matched graphs correspond to the elements of the discrete sets to be matched.
Graph edges define the cost structure of the problem: pairs of matched vertices are penalized in addition to the vertex-to-vertex matchings.
This allows to take, \eg, the underlying geometrical relationship between vertices into account, but also makes the optimization problem NP-hard.

\emph{Deep graph matching}~\cite{rolinek2020deep} is a modern learning-based approach, that combines neural networks for computing matching costs with combinatorial graph matching algorithms to find a matching.
The graph matching algorithm plays a crucial role in this context, as it has to provide high-quality solutions within a limited time budget.
The high demand on run-time is also due to back-propagation learning and graph matching minimization being interleaved and executed together many times during training.

Hence, our work focuses on the \emph{optimization} part of the graph matching pipeline. The modeling and learning aspects are beyond its scope.
We evaluate a range of existing \emph{open-source} algorithms.
Our study compares their performance on a diverse set of computer vision problems.
The focus of the evaluation lies on both, speed and objective value of the solution.

\Paragraph{Why do we require a benchmark?}
Dozens of algorithms addressing the graph matching problem have been proposed in the computer vision literature, see, \eg, the surveys~\cite{foggia2014graph,yan2016short,sun_survey_2020}, and references therein.
Most works promise state-of-the-art performance, which is persuasively demonstrated by experimental evaluation.
However,
\begin{Itemize}
\Item results from one article are often incomparable to results from another, since different problem instances with different costs are used, even if these instances are based on the same image data;
\Item not every existing method is evaluated on all available problem instances, even if open-source code is available.
  Some methods, especially those with poor performance on many instances, are very popular as baselines, whereas better performing techniques are hardly considered in comparisons;
\Item new algorithms are often only evaluated on easy, small-scale problems.
  This does not provide any information on how these algorithms perform on larger, more difficult problem instances.
\end{Itemize}
For these reasons, the field of graph matching has, in our view, not developed as well as it could have done.
By providing a reproducible and extensible benchmark we hope to change this in the future. Such a benchmark is of particular importance for the fast-moving field of \emph{deep graph matching}, as it helps to select an appropriate, fast solver for the combinatorial part of the learning pipeline.

\Paragraph{Graph matching problem.}
Let $\SV$ and $\SL$ be the two finite sets, whose elements we want to match to each other.
For each pair $i,j\in \SV$ and each pair $s,l\in\SL$ a \emph{cost} $c_{is,jl}\in\BR$ is given.
Each pair can be interpreted as an \emph{edge} between a pair of \emph{vertices} of an underlying graph.
This is where the term \emph{graph matching} comes from.
Note that direct \emph{vertex-$i$-to-vertex-$s$} matching costs are defined by the \emph{diagonal} elements $c_{is,is}$ of the resulting \emph{cost} (or \emph{affinity}) \emph{matrix}
$C=(c_{is,jl})$ with $is = (i,s) \in\SV\times\SL$, $jl = (j,l)\in\SV\times\SL$.
The diagonal elements are referred to as \emph{unary costs}, in contrast to the \emph{pairwise costs} defined by non-diagonal entries.
The goal of graph matching is to find a \emph{matching}, or mutual \emph{assignment}, between elements of the sets $\SV$ and $\SL$ that minimizes the total cost for all pairs of assignments.
It can be formulated as the following integer quadratic problem%
\footnote{For sets $A$ and $B$ the notation $x\in A^B$ denotes a vector $x$ whose coordinates take on values from the set $A$ and are indexed by elements of $B$, \ie, each element of $B$ corresponds to a value from $A$.}:
\begin{equation}
  \label{equ:graph-matching-def}
  \begin{gathered}
    \min_{x\in\{0,1\}^{\SV\times \SL }} \sum_{i,j\in\SV} \sum_{s,l\in\SL} c_{is,jl} \, x_{is} \, x_{jl}
    \\
    \quad\text{s.t. }
    \begin{cases}
    \forall i\in\SV\colon\sum_{s\in\SL}x_{is}\le 1\,, \text{ and} \\
    \forall s\in\SL\colon\sum_{i\in\SV}x_{is}\le 1\,.
    \end{cases}
  \end{gathered}
\end{equation}
The vector $x$ defines the matching as $x_{is}=1$ corresponds to assigning $i$ to $s$.
The inequalities in~\eqref{equ:graph-matching-def} allow for this assignment to be \emph{incomplete}, \ie, some elements of both sets may remain unassigned.
This is in contrast to \emph{complete} assignments, where each element of $\SV$ is matched to \emph{exactly one} element of $\SL$ and vice versa.
Note that complete assignments require $|\SV|=|\SL|$.

\Paragraph{Relation to the quadratic assignment problem.}
The graph matching problem~\eqref{equ:graph-matching-def} is closely related to the NP-hard~\cite{Pardalos1993uo} \emph{quadratic assignment problem~(QAP)}~\cite{BurkardQAP}, which is well-studied in operations research~\cite{BurkardQAP,AnalyticalSurveyQAP,cela2013quadratic}.
The QAP only considers \emph{complete} assignments, \ie, $|\SV|=|\SL|$ and equality is required in the constraints in~\eqref{equ:graph-matching-def}.
In contrast, in the field of computer vision incomplete assignments are often required in the model to allow for, \eg, outliers or matching of images with different numbers of features.
Still, graph matching and the QAP are polynomially reducible to each other, see supplement for a proof.

The most famous QAP benchmark is the QAPLIB~\cite{burkard1997qaplib} containing $136$ problem instances.
However, the benchmark problems in computer vision~(CV) substantially differ from those in QAPLIB both by the feasible set that includes incomplete assignments, as well as by the structure of the cost matrix~$C$:
\begin{Itemize}
\Item CV problems are usually of a general, more expressive \emph{Lawler} form~\cite{lawler1963quadratic}, whereas QAPLIB considers factorizable costs $c_{is,jl}=f_{ij}d_{sl}$ known as \emph{Koopmans-Beckmann} form.
  The latter allows for more efficient specialized algorithms.
\Item the cost matrix $C$ in QAPLIB is often dense, whereas in CV problems it is typically sparse, \ie, a large number of entries in $C$ are $0$;
\Item for CV problem instances the cost matrix $C$ may contain \emph{infinite} costs on the diagonal to prohibit certain vertex-to-vertex mappings;
\Item QAPLIB problems are different from an optimization point of view.
  For instance, while the classical LP relaxation~\cite{adams1994improved} is often quite loose for QAPLIB problem instances, it is tight or nearly tight for typical instances considered in CV.
\end{Itemize}

Consequently, comparison results on QAPLIB and CV instances differ significantly.
It is also typical for NP-hard problems that instances coming from different applied areas require different optimization techniques.
Therefore, a dedicated benchmarking on the CV datasets is required.

\Paragraph{Contributions.} Our contribution is three-fold:
\begin{Itemize}
\Item
  Based on open source data, we collected, categorized and generated 451~\emph{existing} graph matching instances grouped into 11~datasets in a common format.
  Most graph matching papers use only a small subset of these datasets for evaluation.
  Our format provides a \emph{ready-to-use} cost matrix $C$ and does not require any image analysis to extract the costs.
\Item
  We collected and categorized 20~open-source graph matching algorithms and evaluated them on the above datasets.
  During that we adapted the cost matrix to requirements of particular algorithms where needed.
  For each method we provide a brief technical description.
  We did not consider algorithms with no publicly available open source code.
\Item
  To allow our benchmark to grow further, we set up a web site%
  \footnote{The web site for the benchmark is available at \url{https://vislearn.github.io/gmbench/}.}
  with all results.
  Our benchmark is reproducible, extensible and follows the best practices of~\cite{beiranvand2017best}.
  We will maintain its web-page in the future and welcome scientists to add problem instances as well as algorithms.
\end{Itemize}

Our work significantly excels evaluations in \emph{all} the papers introducing the algorithms we study.
This implies also to the largest existing comparison~\cite{hutschenreiter_fusionmoves_2021} so far.
The latter considers only 8 out of the 11 datasets and evaluates 6 algorithms out of our 20.

\section{Background to algorithms}
\label{sec:theoretical-preliminaries}
In this section we briefly review the main theoretical concepts and building blocks of the considered approaches.

\Paragraph{Linearization.}
In case all pairwise costs are zero, the objective in~\eqref{equ:graph-matching-def} linearizes to
$\sum_{i\in\SV, s\in\SL} c_{is,is}x_{is}$, turning~\eqref{equ:graph-matching-def} into the \emph{incomplete linear assignment problem~(iLAP)}.
A typical way of how the iLAP is obtained in existing algorithms is by considering the Taylor expansion of the objective~\eqref{equ:graph-matching-def} in the vicinity of a given point~$x$.
The linear term of this expansion forms the iLAP objective.
The iLAP can be reduced to a complete linear assignment problem (LAP)~\cite{burkard2012assignment}, see supplement, and addressed by, \eg, Hungarian~\cite{kuhn1955hungarian} or auction~\cite{bertsekas1979distributed} algorithms.
Below, when we refer to LAP this includes both LAP and iLAP problems.

\Paragraph{Birkhoff polytope and permutation matrices.}
For $|\SV|=|\SL|$ and $x\in[0,1]^{\SV\times\SL}$, where $[0,1]$ denotes the closed interval from $0$ to $1$, the constraints
$\sum_{s\in\SL}x_{is}{=}1$, $\forall i\in\SV$, and $\sum_{i\in\SV}x_{is}{=}1$, $\forall s\in\SL$,
define the set of \emph{doubly-stochastic matrices} also known as \emph{Birkhoff polytope}.
Its restriction to binary vectors $x\in\{0,1\}^{\SV\times\SL}$ is called the \emph{set of permutations} or \emph{permutation matrices}.

\Paragraph{Inequality to equality transformation.}
By adding \emph{slack} or \emph{dummy} variables, indexed by $\#$, with zero cost in the objective in~\eqref{equ:graph-matching-def}, the uniqueness constraints in~\eqref{equ:graph-matching-def} can be rewritten as equalities for $\SV^\#:=\SV \dotcup \{\#\}$, $\SL^\#:=\SL \dotcup \{\#\}$, and $x\in [0,1]^{\SV^{\#}\times\SL^{\#}}$, where $\dotcup$ is the disjoint union:
\begin{equation}
  \label{equ:double-semi-stochastic-set}
  \SB := \left\{
    x \;\middle|\;
    \begin{array}{rl}
      \forall i\in\SV\colon
      & \sum_{s\in\SL^{\#}}x_{is}=1\,, \quad\text{and}
      \\
      \forall s\in\SL\colon
      & \sum_{i\in\SV^{\#}}x_{is}=1
    \end{array}
  \right\}\,.
\end{equation}
Here, $x_{i\#}=1$ (or $x_{\#s}=1$) means that the node $i$ (or label $s$) is unassigned.
Following~\cite{zass_probabilistic_2008}, we refer to the elements of $\SB$ as \emph{doubly-semi-stochastic matrices}.

\Paragraph{Doubly-stochastic relaxation.}
Replacing the \emph{integrality constraints} $x\in\{0,1\}^{\SV\times\SL}$ in ~\eqref{equ:graph-matching-def} with the respective \emph{box constraints} $x\in [0,1]^{\SV\times\SL}$ leads to a \emph{doubly-stochastic relaxation} of the graph matching problem.%
\footnote{%
  Strictly speaking, the term \emph{doubly-stochastic} corresponds to the case when equality constraints are considered in~\eqref{equ:graph-matching-def}.
  In~\cite{zass_probabilistic_2008} the inequality variant is called \emph{doubly semi-stochastic} but we use \emph{doubly-stochastic} in both cases.}
Despite the convexity of its feasible set, the doubly-stochastic relaxation is NP-hard because of the non-convexity of its quadratic objective in general~\cite{sahni1974}.

\Paragraph{Probabilistic interpretation.}
Doubly-stochastic relaxations are often motivated from a probabilistic perspective, where the individual matrix entries represent matching probabilities.
An alternative probabilistic interpretation is to consider the \emph{product graph} between $\SV$ and $\SL$, in which the edge weights directly depend on the cost matrix $C$.
This way, graph matching can be understood as selecting reliable nodes in the product graph, \eg, by random walks~\cite{RandomWalksForGraphMatching}.

\Paragraph{Injective and bijective formulations.}
Assume $|\SV|\le|\SL|$.
A number of existing approaches consider an asymmetric formulation of the graph matching problem~\eqref{equ:graph-matching-def}, where the uppermost constraint in~\eqref{equ:graph-matching-def} is exchanged for equality, \ie, $\forall {i\in\SV}\colon\sum_{s\in\SL}x_{is}=1$. We call this formulation \emph{injective}. The strict inequality case $|\SV| < |\SL|$ is also referred to as an \emph{unbalanced QAP} in the literature. Note that to address problems of the general form~\eqref{equ:graph-matching-def} by such algorithms, it is necessary to extend the set $\SL$ with $|\SV|$ dummy elements.
This is similar to the reduction from graph matching to QAP described in the supplement.
Since availiable implementations of multiple considered algorithms are additionally restricted to the case $|\SV|=|\SL|$, \ie to the classical QAP as introduced in~\cref{sec:intro}, we adopt the term \emph{bijective} to describe the corresponding algorithms and datasets.

\Paragraph{Spectral relaxation.}
The graph matching objective in~\eqref{equ:graph-matching-def} can be compactly written as $\vecx^\top C \vecx$.
Instead of the uniqueness constraints in~\eqref{equ:graph-matching-def} the \emph{spectral relaxation} considers the non-convex constraint $\vecx^\top \vecx = n$.
This constraint includes all matchings with exactly $n$ assignments, which is of interest when the total number of assignments $n$ is known, \eg, for the injective formulation where $n=|\SV|$.
The minimization of $\vecx^\top C \vecx$ subject to $\vecx^\top \vecx = n$ reduces to an \emph{eigenvector problem}, \ie, finding a vector $\vecx$ corresponding to the smallest eigenvalue of the matrix $C$.
The latter amounts to minimizing a Rayleigh quotient~\cite{horn2012matrix}.

\Paragraph{Path following.}
Another way to deal with the non-convexity of the graph matching problem is \emph{path-following}, represented by~\cite{FactorizedGraphMatching} in our study.
The idea is to solve a sequence of optimization problems with objective $f_{\alpha^t}(\vecx) = (1-\alpha^t) f_{\text{cvx}}(\vecx) + \alpha^t f_{cav}(\vecx)$ for $\alpha^t$, $t\in 1,\dots, N$, gradually growing from $0$ to $1$.
The (approximate) solution of each problem in the sequence is used as a starting point for the next.
The hope is that this iterative process, referred to as \emph{following the convex-to-concave path}, leads to a solution with low objective value for the whole problem.
The objective for $\alpha^1=0$ is equal to $f_{\text{cvx}}(\vecx)$ and is convex, therefore it can be solved to global optimality.
The objective for $\alpha^N=1$ is equal to $f_{\text{cav}}(\vecx)$ and is concave.
Its local optima over the set of doubly-stochastic matrices are guaranteed to be binary, and, therefore, \emph{feasible assignments}, \ie, they satisfy all constraints of~\eqref{equ:graph-matching-def}.

\Paragraph{Graphical model representation.}
The graph matching problem can be represented in the form of a \emph{maximum a posteriori (MAP) inference} problem for discrete graphical models~\cite{savchynskyy2019discrete}, known also as \emph{Markov random field (MRF) energy minimization} and closely related to \emph{valued and weighted constraint satisfaction} problems.
As several graph matching works in computer vision~\cite{HungarianBP,swoboda2017study,hutschenreiter_fusionmoves_2021}, use this representation, we provide it below in more detail.

Let $(\SV,\SE)$ be an undirected graph, with the finite set $\SV$ introduced above being the \emph{set of nodes} and $\SE\subseteq\binom{\SV}{2}$ being the set of \emph{edges}.
For convenience we denote edges $\{i,j\}\in \SE$ simply by $ij$.
Let the finite set $\SL$ introduced above be the set of \emph{labels}.
We associate with each node $i\in \SV$ a set  $\mathcal{L}^{\#}_i=\mathcal{L}_i \dotcup \{\#\}$ with $\mathcal{L}_i\subseteq\mathcal{L}$.
Like above, $\#$ stands for the \emph{dummy label} distinct from all labels in $\SL$ to encode that \emph{no label} is selected.
With each label $s\in\SL^{\#}_i$ in each node $i\in\SV$ the \emph{unary cost} $\theta_{is}:=c_{is,is}$ ($0$ for $s=\#$) is associated.
The case $|\SL_i| < |\SL|$ corresponds to infinite unary costs $c_{is,is}=\infty$, $s\in\SL\backslash \SL_i$, as the respective assignments can be excluded from the very beginning.
Likewise, with each edge $ij\in \SE$ and each label pair $sl\in \SL_i^{\#}\times\SL_j^{\#}$, the pairwise cost $\theta_{is,jl}=c_{is,jl}+c_{jl,is}$ ($0$ for $s$ or $l=\#$) is associated.
The graph $(\SV,\SE)$ being undirected implies $ij=ji$ and $\theta_{is,jl}=\theta_{jl,is}$.
An edge $ij$ belongs to $\SE$ only if there is a label pair $sl\in \SL_i\times\SL_j$ such that $\theta_{is,jl}\neq 0$. In this way a sparse cost matrix $C$ may translate into a sparse graph~$(\SV,\SE)$.

The problem of finding an optimal assignment of labels to nodes, equivalent to the \emph{graph matching} problem~\eqref{equ:graph-matching-def}, can thus be stated as
\begin{equation}
  \label{equ:map-formulation}
  \begin{gathered}
    \min_{y\in \SY} \Big[E(y):= \sum_{i\in \SV}\theta_{iy_i} + \sum_{ij \in \SE}\theta_{iy_i,jy_j}\Big]
    \\
    \text{s.\,t.}\quad \forall\, i,j\in \SV, i\neq j\colon y_i \neq y_j \text{ or } y_i = \#
  \end{gathered}
\end{equation}
where $\SY$ stands for the Cartesian product $\bigtimes _{i\in \SV} \SL^\#_i$, and $y_i=s$, $s\in\SL_i$, is equivalent to $x_{is}=1$ in terms of~\eqref{equ:graph-matching-def}.
Essentially,~\eqref{equ:map-formulation} corresponds to a \emph{MAP inference problem for discrete graphical models}~\cite{savchynskyy2019discrete} with additional uniqueness constraints for the labels.

\Paragraph{ILP representation and LP relaxations.}
Based on~\eqref{equ:map-formulation} the graph matching problem can be expressed by a linear objective subject to linear and integrality constraints by introducing variables $x_{is,jl}= x_{jl,is}$ for each pair of labels $sl\in\SL_i^\#\times\SL_j^\#$ in neighboring nodes $ij\in\SE$, and enforcing the equality $x_{is,jl}=x_{is}x_{jl}$ with suitable linear constraints.
An \emph{integer linear program (ILP)} formulation of the graph matching problem~\eqref{equ:graph-matching-def} can then be written as:
\begin{gather}
  \label{equ:ILP-representation}
  \min_{x\in\{0,1\}^\SJ} \sum_{\substack{i\in \SV \\ s\in\SL^\#_i}} c_{is}x_{is} + \sum_{\substack{ij \in \SE \\ \mathclap{sl\in\SL^\#_i\times\SL^\#_j}}} (c_{is,jl}+c_{jl,is}) \, x_{is,jl}\\
  \forall i\in\SV\colon \sum_{s\in\SL^\#_i}x_{is}=1\,,\hspace{10pt}%
  \forall s\in\SL\colon \sum_{i\in\SV}x_{is}\le 1\,,\label{equ:uniqueness-constraints-ILP}\\
  \forall ij\in\SE,\ l\in\SL^\#_j\colon \sum_{s\in\SL^\#_i}x_{is,jl}=x_{jl}\,. \label{equ:coupling-constraints-ILP}
\end{gather}
Here $\SJ=\{(i,s)\colon i\in\SV,\ s\in\SL^\#_i\}\cup\{(is,jl)\colon ij\in\SE, sl\in\SL^\#_i\times\SL^\#_j\}$ denotes the set of coordinates of the vector $x$.
The formulation~\eqref{equ:ILP-representation}-\eqref{equ:coupling-constraints-ILP} differs from the standard ILP representation for discrete graphical models by the label uniqueness constraints~(\ref{equ:uniqueness-constraints-ILP}, rightmost).
Substitution of the integrality constraints $x\in\{0,1\}^\SJ$ in~\eqref{equ:ILP-representation} with the box constraints $x\in[0,1]^\SJ$ results in the respective \emph{LP relaxation}.

\input{floats/table-methods}

\section{Graph matching algorithms}

Below we summarize the graph matching methods that we consider in our comparison, see \cref{tab:methods} for an overview of their characteristics and references.

\subsection{Primal heuristics}

\Paragraph{Linearization based.}
These methods are based on iterative linearizations of the quadratic objective~\eqref{equ:graph-matching-def} derived from its Taylor expansion.

\emph{Iterated projected fixed point} (\Salg{ipfp})~\cite{IntegerFixedPointGraphMatching} solves on each iteration the LAP obtained through linearization in the vicinity of a current, in general non-integer, assignment.
Between iterations the quadratic objective is optimized along the direction to the obtained LAP solution, which yields a new, in general non-integer assignment.
We evaluate two versions of \Salg{ipfp} which differ by their initialization:
\Salg{ipfpu} is initialized with $x^0\in [0,1]^{\SV\times\SL}$, where $x^0_{is}=1/\sqrt{N}$ if $c_{is,is}<\infty$, and $x^0_{is}=0$ otherwise.
Here, $N:= |\{ is\in\SV\times\SL \mid c_{is,is}< \infty \}|$.
\Salg{ipfps} starts from the result of the spectral matching \Salg{sm}~\cite{leordeanu_spectral_2005} described below.

\emph{Graduated assignment} (\Salg{ga})~\cite{GraduatedAssignmentGold} optimizes the doubly-stochastic relaxation.
On each iteration it approximately solves the LAP obtained through linearization in the vicinity of a current, in general non-integer, assignment utilizing the Sinkhorn algorithm~\cite{kosowsky_invisible_1994} for a given fixed temperature.
The obtained approximate solution is used afterwards as the new assignment.
The temperature is decreased over iterations to gradually make the solutions closer to integral.

\emph{Fast approximate quadratic programming} (\Salg{fw})~\cite{vogelstein_fast_2015} considers the Frank-Wolfe method~\cite{frank1956algorithm} for optimizing over the set $\SB$, \cf~\eqref{equ:double-semi-stochastic-set}.
Each iteration first solves a LAP to find the optimum of the linearization at the current solution, followed by a line search in order to find the best convex combination of the current and the new solution.
To obtain an integer solution, the objective of the LAP solution is evaluated in each iteration, and the lowest one among all solutions is kept.
The initial LAP is based on the unary costs only.
The implementation~\cite{code_swoboda} we evaluate is applicable to the general Lawler form of the problem~\eqref{equ:graph-matching-def}, in contrast to the Koopmans-Beckmann form addressed in~\cite{vogelstein_fast_2015}.

\Paragraph{Norm constraints based.}
\emph{Spectral matching} (\Salg{sm})~\cite{leordeanu_spectral_2005}
uses a spectral relaxation that amounts to a Rayleigh quotient problem~\cite{horn2012matrix} which can be optimized by the power iteration method. Here, each update comprises of a simple matrix multiplication and a subsequent normalization, so that $\vecx^t$ is iteratively updated via $\vecx^{t+1} =  -C\vecx^t / \|  C\vecx^t\|_2$.

\emph{Spectral matching with affine constraints} (\Salg{smac})~\cite{cour2007balanced}
is similar to \Salg{sm}, but additionally takes into account affine equality constraints that enforce one-to-one matchings. The resulting formulation  amounts to a Rayleigh quotient problem under affine constraints, that can efficiently be computed in terms of the eigenvalue decomposition.

\emph{Max-pooling matching} (\Salg{mpm})~\cite{cho_finding_2014} resembles \Salg{sm}, but it replaces the sum-pooling implemented in terms of the matrix multiplication $-C \vecx$ in the power iteration update of \Salg{SM} by a max-pooling operation.
With that, only candidate matches with the smallest costs are taken into account.

\emph{Local sparse model} (\Salg{lsm})~\cite{jiang2015local} solves the relaxation
$\max_{x} x^\top C x$, s.t.~$||x||^2_{1,2} = \sum_{i=1}^{|\SV|} \big( \sum_{k=1}^{|\SL|} |x_{ik}| \big)^2 = 1$, $x \geq 0$.
The $l_{1,2}$-norm $||x||_{1,2}$ should encourage the solution of the above relaxation to be sparse in each row when treating $x$ as a matrix.
This resembles the sparsity property of permutation matrices, which satisfy $||x||_{1,2}=|\SV|$.

\begin{remark}
  \label{rem:non-pos}
  All of the norm constraints based algorithms described above require non-positive%
  \footnote{Non-negative in original maximization formulations}
  costs in order to guarantee convergence of the underlying iterative techniques.
  This condition can be w.l.o.g.\ assumed for any graph matching problem.
  The corresponding cost transformation is described in the supplement.
\end{remark}

\Paragraph{Probabilistic interpretation based.}
\emph{Reweighted Random Walks Matching}~(\Salg{rrwm})~\cite{RandomWalksForGraphMatching} interprets graph matching as the problem of selecting reliable nodes in an \emph{association graph}, whose weighted adjacency matrix is given by $-C$.
Nodes are selected through a random walk that starts from one node and randomly visits nodes according to a Markov transition matrix derived from the edge weights of the association graph.
In order to take into account matching constraints, the authors of~\cite{RandomWalksForGraphMatching} consider a reweighted random walk strategy.

\emph{Probabilistic matching}~(\Salg{pm})~\cite{zass_probabilistic_2008} considers a probabilistic formulation of graph matching in which the quadratic objective is replaced by a relative entropy objective.
It is shown that by doing so one can obtain a convex problem formulation via marginalization, which is optimized in terms of an iterative successive projection algorithm.

\begin{remark}
  \label{rem:discretization}
  Most of the primal heuristics considered above aim to optimize the quadratic objective~\eqref{equ:graph-matching-def} over a continuous set such as, \eg, the Birkhoff polytope.
  The resulting assignment $x\in\BR^{\SV\times\SL}$ is, therefore, not guaranteed to be integer.
  As suggested in~\cite{RandomWalksForGraphMatching}, to obtain an integer assignment we solve a LAP with $(-x)$ treated as the cost matrix.
  We apply this procedure as a postprocessing step for \Salg{ipfp}, \Salg{ga}, \Salg{sm}, \Salg{smac}, \Salg{mpm}, \Salg{lsm}, \Salg{rrwm}, and \Salg{pm}.
  Note that this postprocessing does not change an integer assignment.
\end{remark}

\Paragraph{Path following based.}
\emph{Factorized graph matching} (\Salg{fgmd})~\cite{FactorizedGraphMatching} proposes an efficient factorization of the cost matrix to speed-up computations, and is based on the convex-concave path following strategy, see~\cref{sec:theoretical-preliminaries}.
Individual problems from the path are solved with the Frank-Wolfe method~\cite{frank1956algorithm}.

\input{floats/table-bca-algorithms}

\Paragraph{Randomized generation and fusion based.}
\emph{Fusion moves with a greedy heuristic} (\Salg{fm})~\cite{hutschenreiter_fusionmoves_2021}
is based on the graphical model representation and consists of two parts: A randomized greedy assignment generation, and \emph{fusion} of the assignments.
The randomized generator greedily fixes labels in the nodes in a way that minimizes the objective value restricted to the already fixed labels.
The fusion procedure merges the current assignment with the next generated one by approximately solving an auxiliary \emph{binary} MAP inference problem utilizing QPBO-I~\cite{rother07-cvpr}.
The merged solution is guaranteed to be at least as good as the two input assignments.
This property guarantees monotonic improvement of the objective value.

\subsection{Lagrange duality-based techniques}

The methods below consider the Lagrange decompositions~\cite{guignard1987lagrangean} of the graph matching problem~\eqref{equ:graph-matching-def}~\cite{GraphMatchingDDTorresaniEtAl},
or its graphical model representation~\eqref{equ:map-formulation}~\cite{HungarianBP,swoboda2017study,hutschenreiter_fusionmoves_2021}, and optimize the corresponding dual.
The methods differ in the dual optimization and chosen primal solution reconstruction algorithms.

\Paragraph{Block-coordinate methods (\Salg{hbp}, \Salg{mp-*}, \Salg{fm-bca}).}
The works~\cite{HungarianBP,swoboda2017study,hutschenreiter_fusionmoves_2021} employ a block-coordinate ascent (BCA) technique to optimize the dual problem obtained by relaxing the coupling~\eqref{equ:coupling-constraints-ILP} and label uniqueness constraints~(\ref{equ:uniqueness-constraints-ILP}, rightmost).
Since the dual is piece-wise linear, BCA algorithms may not attain the dual optimum, but may get stuck in a sub-optimal fixed point~\cite{bertsekas1999nonlinear,savchynskyy2019discrete}.

Although the elementary operations performed by these algorithms are very similar, their convergence speed and attained fixed points differ drastically.
In a nutshell, these methods decompose the problem~\eqref{equ:map-formulation} into the graphical model without uniqueness constraints, and the LAP problem, as described, \eg, in~\cite{hutschenreiter_fusionmoves_2021}.
Dual algorithms reparametrize the problem making it more amenable to primal techniques~\cite{savchynskyy2019discrete}.
\Cref{tab:bca-algorithms} gives an overview of the evaluated combinations for
\begin{Itemize}
\Item optimizing the dual of the graphical model,
\Item optimizing the LAP, and
\Item obtaining the primal solution from the reparametrized costs%
\footnote{\emph{Reparametrized} costs are also known as \emph{reduced} costs, \eg, in the simplex tableau.},
which influence the practical performance of BCA solvers.
\end{Itemize}
Additionally, \Salg{mp-mcf} and \Salg{mp-fw} tighten the relaxation by considering triples of graph nodes as subproblems.

\Paragraph{Subgradient method (\Salg{dd-ls*}).}
The algorithms denoted as \Salg{dd-ls*} with \Salg{*} being \Salg{0}, \Salg{3} or \Salg{4} represent different variants of a dual subgradient optimization method~\cite{GraphMatchingDDTorresaniEtAl}.
The variant \Salg{dd-ls0} addresses the relaxation equivalent to a symmetrized graphical model formulation, see supplement for a description.
This is achieved by considering the Lagrange decomposition of the problem into two graphical models, with $\SV$ and $\SL$ being the set of nodes, respectively, and a LAP subproblem.
The graphical models are further decomposed into acyclic ones, \ie~trees, solvable by dynamic programming, see, \eg,~\cite[Ch.9]{savchynskyy2019discrete}.
The \emph{tree decomposition} is not described in~\cite{GraphMatchingDDTorresaniEtAl}, and we reconstructed it based on the source code~\cite{code_kolmogorov} and communication with the authors.
As we observed it to be more efficient than the \emph{max-flow subproblems} suggested in the paper~\cite{GraphMatchingDDTorresaniEtAl} the latter were not used in our evaluation.

Variants \Salg{dd-ls3} and \Salg{dd-ls4} tighten the relaxation of \Salg{dd-ls0} by considering \emph{local subproblems} of both graphical models in the decomposition.
These are obtained by reducing the node sets $\SV$ and $\SL$ to $3$ or respectively $4$ elements inducing a connected subgraph of the graphical model, see~\cite{GraphMatchingDDTorresaniEtAl} for details.

\section{Benchmark}

\Paragraph{Datasets.}
The 11 datasets we collected for evaluation of the graph matching algorithms stem from applications in computer vision and bio-imaging.
All existing graph matching papers use only a subset of these datasets for evaluation purposes.
Together these datasets contain 451 problem instances.
\Cref{tab:datasets} gives an overview of their characteristics.
We modified costs in several datasets to make them amenable to some algorithms, see supplement.
Our modification results in a constant shift of the objective value for each feasible assignment, and, therefore, does not influence the quality of the solution.

\input{floats/table-datasets}

Below we give a brief description of each dataset.
Along with the standard computer vision datasets with small-sized problems, \Sdata{hotel}, \Sdata{house-dense/sparse}, \Sdata{car}, \Sdata{motor} and \Sdata{opengm} with $|\SV|$ up to $52$, our collection contains the middle-sized problems \Sdata{flow}, with $|\SV|$ up to $126$, and the large-scale \Sdata{worms} and \Sdata{pairs} problems with $|\SV|$ up to 565.

\emph{Wide baseline matching} (\Sdata{hotel}, \Sdata{house-dense/sparse})
is based on a series of images of the same synthetic object with manually selected landmarks from different viewing angles based on the work by~\cite{CaetanoMCLS09}.
For \Sdata{hotel} and \Sdata{house-dense} we use the same models as in~\cite{swoboda2017study} published in~\cite{house_hotel_models}.
\Sdata{house-sparse} consists of the same image pairs as \Salg{house-dense}, but the cost structure is derived following the approach of~\cite{HungarianBP} that results in significantly sparser problem instances.
Graphs with the landmarks as nodes are obtained by Delaunay triangulation.
The costs are set to
$c_{is,jl}=-\exp(-(d_{ij}-d_{sl})^2/2500)A^1_{ij}A^2_{sl}$ where $d_{ij},d_{sl}$
are Euclidean distances between two landmarks and
$A^1\in\{0,1\}^{\SV\times \SV}$, $A^2\in\{0,1\}^{\SL\times \SL}$
are adjacency matrices of the corresponding graphs.
The unary costs are zero.

\emph{Keypoint matching} (\Sdata{car}, \Sdata{motor})
contains \emph{car} and \emph{motor}bike images from the PASCAL VOC 2007 Challenge~\cite{VocPascal} with the features and costs from~\cite{UnsupervisedLearningForGraphMatching}.
We use the instances available from~\cite{fusionMovesProjectPage}.

\emph{Large displacement flow} (\Sdata{flow})
was introduced by~\cite{GraphFlow} for key point matching on scenes with large motion.
We use the instances from~\cite{fusionMovesProjectPage} which use keypoints and costs as in~\cite{swoboda2017study}.

\emph{OpenGM matching} (\Sdata{opengm})
is a set of non-rigid point matching problems by~\cite{KomodakisP08}, now part of the \emph{OpenGM} Benchmark~\cite{OpenGMBenchmark}.
We use the instances from~\cite{fusionMovesProjectPage}.

The \Sdata{caltech} dataset was proposed in \cite{RandomWalksForGraphMatching}.
The data available at the project page~\cite{caltech} contains the \emph{mutual projection error} matrix $D=(d_{is,jl})$, lists of possible assignments, and partial ground truth.
We reconstructed the dataset from this data.
Unary costs are set to zero. Pairwise costs for pairs of possible assignments are set to
$c_{is,jl} = -\max (50 - d_{is,jl},0)$.
We divided the dataset into \Sdata{caltech-small} and \Sdata{caltech-large}, where all instances with more than $40000$ non-zero pairwise costs are considered as large.

\emph{Worm atlas matching} (\Sdata{worms})
has the goal to annotate nuclei of \emph{C.~elegans}, a famous model organism used in developmental biology, by assigning nuclei names from a known atlas of the organism.
A detailed description can be found in~\cite{kainmueller2014active}.
We use the instances obtained from~\cite{fusionMovesProjectPage} which are originally from~\cite{kainmueller2017graph}.

\emph{Worm-to-worm matching} (\Sdata{pairs})
directly matches the cell nuclei of individual \emph{C.~elegans} worms to each other.
The resulting models are much coarser than those of the \Sdata{worms} dataset.
We consider the same 16 problem instances as~\cite{hutschenreiter_fusionmoves_2021} using the models from~\cite{fusionMovesProjectPage}.

\Paragraph{Evaluation metrics.}
For \emph{fixed-time} performance evaluation~\cite{beiranvand2017best} we restrict run-time (1, 10, 100\,s) and evaluate attained objective values \emph{E}, lower bound \emph{D} and, for datasets with ground truth available, accuracy \emph{acc}.
We also report the number of optimally solved instances per dataset.

For \emph{fixed-target} performance evaluation~\cite{beiranvand2017best} we measure the time $t_s(p)$ until each solver $s$ solves the problem $p$ within an optimality tolerance of 0.1\%.
For instances with unknown optimum, we consider the best achieved objective value across all methods as optimum as suggested in~\cite{beiranvand2017best}.
The performance ratio to the best solver is computed by $r_s(p) = \frac{t_s(p)}{\min \{ t_s(p) : \forall s \}}$.
We create a performance profile~\cite{beiranvand2017best,dolan2002benchmarking} by computing $\rho_s(\tau) = \frac{1}{|P|} \cdot |\{ r_s(p) \leq \tau : \forall p \}|$ for each solver $s$ where $|P|$ denotes the total number of problem instances.
Intuitively, $\rho_s(\tau)$ is the probability of solver $s$ being at most $\tau$ times slower than the fastest solver.

\input{floats/figure-performance-profile}

\section{Empirical Results}

Fixed-time evaluation presented in \cref{tab:comparison-small-datasets} addresses small problem instances, whereas \cref{tab:comparison-large-datasets} addresses mid-size and large problem instances.
The performance profile for fixed-target evaluation is presented in \cref{fig:performance-profile}.
More detailed results are available in the supplement.
Results have been obtained by taking the minimum run-time across five trials on an AMD EPYC 7702 2.0\,GHz processor.
Randomized alogrithms were made deterministic by fixing their random seed (\Salg{fm} and \Salg{fm-bca}).
We equally treat Matlab and \Cpp implementations, in spite of the apparent efficiency considerations.
The reason for this is that the solution quality of \emph{all} Matlab algorithms in our evaluation is inferior to the \Cpp techniques, even if run-time is ignored.

\input{floats/table-small-problems}
\input{floats/table-large-problems}

For \textbf{small problems} we show results for $1$ second in \cref{tab:comparison-small-datasets}, as the best methods already solve almost all instances to optimality within this time.
The best methods on these datasets are \Salg{fm}, \Salg{fm-bca} and \Salg{dd-ls0}.
\Salg{dd-ls3/4} have higher costs per iteration, and require more than 1 second to arrive at the solution quality of \Salg{dd-ls0}.
The other dual BCA-based methods perform almost as good on all but the \Sdata{opengm} dataset, which seems to be the most difficult dataset amongs the one in \cref{tab:comparison-small-datasets}.
Apart from \Salg{fm} pure primal heuristics are unable to compete with duality-based techniques.
The comparison of the results for \Sdata{house-dense} and \Sdata{house-sparse} shows that most of the primal heuristics perform much better on sparse problems.

For \textbf{larger problems} the most representative times shown in \cref{tab:comparison-large-datasets} are 1, 10 and 100 seconds, depending on the dataset.
Again, the duality-based methods and the \Salg{fm} heuristic lead the table.
The \Salg{fm-bca} method consistently attains the best or close to best objective and accuracy values on all datasets, whereas its lower bound is often worse than the lower bounds obtained by the \Salg{mp-*} and \Salg{dd-ls*} methods.
In contrast, most of the primal heuristics as well as \Salg{hbp} fail, and, for brevity, are omitted in \cref{tab:comparison-large-datasets}.

Algorithms \Salg{dd-ls3}/\Salg{4} consider tighter relaxations than \Salg{dd-ls0}, but are slower, therefore lose in the competition on short time intervals.
However, they have the ability to attain the best lower bounds given longer runs ($\gg$ 100s).

There is a significant performance gap between the closely related \Salg{hbp}, \Salg{mp-*} and \Salg{fm-bca} methods.
Foremost, this is explained by the method for reconstructing the primal solution: The \Salg{fm} algorithm used in the \Salg{fm-bca} solver is solid also as a stand-alone technique, and significantly outperforms the \Salg{fw} and LAP heuristics used in the \Salg{mp-*} algorithms.
The branch-and-bound solver used in \Salg{hbp} is quite slow and does not scale well.
The second reason for different performance of these methods is the specific BCA algorithm used for the underlying discrete graphical model, \cf~\cref{tab:bca-algorithms}.
According to the recent study~\cite{tourani2020taxonomy}, which provides a unified treatment of the dual BCA methods for dense%
\footnote{Most of the considered graphical models are dense in terms of~\cite{tourani2020taxonomy}.}
graphical models, MPLP\Rplus\Rplus\ performs best, followed by anisotropic diffusion and MPLP as the slowest method.
\Cref{tab:comparison-large-datasets} shows that there is no solution suitable for every purpose:
The speed of \Salg{fm-bca} comes at the price of a looser lower bound.
Nonetheless, combining a primal heuristic with a dual optimizer consistently improves upon the results obtained by the heuristic alone.
This holds for \Salg{fm}, but the effect is even more pronounced for the \Salg{fw} and LAP heuristics.

The fixed-target evaluation in \cref{fig:performance-profile} confirms that the \Salg{fm} and \Salg{fm-bca} method are amongst the best performing solvers.
While \Salg{fm-bca} uses \Salg{fm} as primal heuristics with additional dual BCA updates, the overhead of the latter is visible.
After increasing the allowed performance ratio for \Salg{fm-bca} to a factor of 3.7, we can expect better solutions than \Salg{fm} alone.
Other top performers are duality-based algorithms with \Salg{mp-fw} and \Salg{dd-ls0} being the closest followers.

\section{Conclusions}

Our evaluation shows that:
\begin{Itemize}
\Item
  Most instances from the popular datasets \Sdata{hotel}, \Sdata{house}, \Sdata{car} and \Salg{motor} can be solved to optimality in well below a second by several optimization techniques.
  \Sdata{opengm} can also be solved to optimality in under a second, although it turns out to be hard for many methods.
  Therefore, we argue that \emph{these datasets alone are not sufficient anymore to empirically show efficiency of new algorithms}.
  The most difficult in our collection are the datasets \Sdata{caltech-*} and \Sdata{pairs}.
  For a comprehensive evaluation of new methods more datasets are required.
\Item
  The most popular comparison baselines like \Salg{ipfp}, \Salg{ga}, \Salg{rrwm}, \Salg{pm}, \Salg{sm}, \Salg{smac}, \Salg{lsm}, \Salg{mpm} and \Salg{fgmd} are not competitive, and, therefore, \emph{comparison to these alone should not anymore be considered as sufficient}.
\Item
  The most efficient methods are duality-based techniques equipped with efficient primal heuristics.
  In particular, the \Salg{fm}/\Salg{fm-bca} method currently attains the best or nearly best objective values for most problem instances in the shortest time.
\Item
  Although being NP-hard in general, the graph matching problem can be often efficiently solved in computer vision practice.
  For many of the considered datasets, including those with $|\SL| > 1000$ and $|\SV|>500$, a reasonable approximate solution can be attained in less than a second.
\end{Itemize}

\section{Acknowlegments}

This work was supported by the German Research Foundation~(Unsupervised Model Discovery for Stereotypical Organisms, DFG SA 2640/2-1)
and the Helmholtz Information \& Data Science School for Health~(HIDSS4Health).
We thank the Center for Information Services and High Performance Computing~(ZIH) at TU Dresden for its facilities for high throughput calculations.
\clearpage
\bibliographystyle{plain}
\bibliography{references}

\clearpage
\input{appendix}

\end{document}


%% file: preamble.tex
\usepackage[utf8]{inputenc}
\usepackage[T1]{fontenc}
\usepackage[english]{babel}

\usepackage{amsmath}
\usepackage{amsthm}
\usepackage{mathtools}

\usepackage{graphicx}
\usepackage[dvipsnames,table]{xcolor}
\usepackage{tikz}

\usetikzlibrary{fit,calc,shadings}

\usepackage{booktabs}
\usepackage{tabularx}
\usepackage{bigdelim}

\newcolumntype{g}{>{\columncolor{gray!25}}c} 

\usepackage[nocompress]{cite}
\usepackage[a4paper, margin=15mm, tmargin=25mm, bmargin=27mm]{geometry}
\usepackage{ragged2e}
\usepackage{relsize}
\usepackage{scrlayer-scrpage}
\usepackage{xspace}
\usepackage[breaklinks=true,colorlinks=true,bookmarks=false,linkcolor=BrickRed,citecolor=Green,hypertexnames=false]{hyperref}

\usepackage[ruled,vlined,nofillcomment]{algorithm2e}
\def\MyAlgBugFix{\addtolength\hsize{1.5em}}

\SetCommentSty{MyCommentStyle}
\SetProgSty{}
\SetAlCapHSkip{0pt}
\let\textvtt\texttt
\let\vttfamily\ttfamily

\usepackage[capitalize,noabbrev]{cleveref}

\makeatletter
\newcommand\Paragraph{\@startsection{paragraph}{4}{\z@}%
                                    {0pt plus .2em}%
                                    {-.4em plus -1.0em}%
                                    {\normalfont\normalsize\bfseries}}
\newcommand\ParagraphRun{\@startsection{paragraph}{4}{\z@}%
                                       {0pt plus .2em}%
                                       {-\fontdimen2\font plus -\fontdimen3\font minus -\fontdimen4\font}%
                                       {\normalfont\normalsize\bfseries}}
\makeatother

\newenvironment{Itemize}{\setcounter{enumi}{0}\ignorespaces}{\ignorespacesafterend}
\newcommand\Item{\noindent\textbf{(\stepcounter{enumi}\roman{enumi})}~}

\addtokomafont{captionlabel}{\bfseries}
\addtokomafont{disposition}{\rmfamily}
\setcapindent{0pt}
\deffootnote{.6em}{2em}{%
	\makebox[.6em][l]{\textsuperscript{\thefootnotemark}}%
}

\usepackage{newpxtext}
\usepackage{newpxmath}

%% file: notation.tex
\newtheorem{theorem}{Theorem}
\newtheorem{corollary}{Corollary}
\newtheorem{remark}{Remark}

\DeclareMathOperator*\argmin{arg\,min}

\newcommand\SB{\mathcal{B}}
\newcommand\SV{\mathcal{V}}
\newcommand\SE{\mathcal{E}}
\newcommand\SJ{\mathcal{J}}
\newcommand\SL{\mathcal{L}}

\newcommand\SY{\mathcal{Y}}
\newcommand\BR{\mathbb{R}}

\newcommand\dotcup{\mathbin{\dot\cup}}

\newcommand\vecx{x}
\newcommand\Salg[1]{\textvtt{#1}}
\newcommand\Sdata[1]{\textvtt{#1}}

\newcommand\eg{e.g.\xspace}
\newcommand\ie{i.e.\xspace}
\newcommand\cf{c.f.\xspace}
\newcommand\wrt{wrt.\xspace}

\newcommand\Rplus{\protect\hspace{-.1em}\protect\raisebox{.35ex}{\smaller{\smaller\textbf{+}}}}
\newcommand\Cpp{\mbox{C\Rplus\Rplus}\xspace}

%% file: floats/table-methods.tex
\begin{table*}[t]
	\small\sffamily
	\setlength{\aboverulesep}{0pt}%
	\setlength{\belowrulesep}{0pt}%
	\setlength\tabcolsep{3pt}%
	\begin{tabular}{l *{8}{gc} g}
		\toprule
		method
			& \rotatebox{90}{IQP}
			& \rotatebox{90}{ILP}
			& \rotatebox{90}{bijective}
			& \rotatebox{90}{non-pos.~}
			& \rotatebox{90}{0-unary}
			& \rotatebox{90}{lineariz.}
			& \rotatebox{90}{norm}
			& \rotatebox{90}{doubly}
			& \rotatebox{90}{spectral}
			& \rotatebox{90}{discret.}
			& \rotatebox{90}{path fol.}
			& \rotatebox{90}{fusion}
			& \rotatebox{90}{duality}
			& \rotatebox{90}{SGA}
			& \rotatebox{90}{BCA}
			& \rotatebox{90}{Matlab}
			& \rotatebox{90}{\Cpp}
			\\ \midrule
		\Salg{fgmd}~\cite{FactorizedGraphMatching}               & + &   & + &   &   &   &   &   &    &   & + &   &   &   &   & \hskip 1sp \cite{code_FGMD}      &                                          \\
		\Salg{fm}~\cite{hutschenreiter_fusionmoves_2021}         &   & + &   &   &   &   &   &   &    &   &   & + &   &   &   &                                  & \hskip 1sp \cite{fusionMovesProjectPage} \\
		\Salg{fw}~\cite{vogelstein_fast_2015}                    & + &   &   &   &   & + &   & + &    &   &   &   &   &   &   &                                  & \hskip 1sp \cite{code_swoboda}           \\
		\Salg{ga}~\cite{GraduatedAssignmentGold}                 & + &   & + &   &   & + &   & + &    & + &   &   &   &   &   & \hskip 1sp \cite{code_cour}      &                                          \\
		\Salg{ipfps}~\cite{IntegerFixedPointGraphMatching}       & + &   & + & + &   & + &   & + &    & + &   &   &   &   &   & \hskip 1sp \cite{code_leordeanu} &                                          \\
		\Salg{ipfpu}~\cite{IntegerFixedPointGraphMatching}       & + &   & + &   &   & + &   & + &    & + &   &   &   &   &   & \hskip 1sp \cite{code_leordeanu} &                                          \\
		\Salg{lsm}~\cite{jiang2015local}                         & + &   & + & + &   &   & + &   &    & + &   &   &   &   &   & \hskip 1sp \cite{code_zhang}     &                                          \\
		\Salg{mpm}~\cite{cho_finding_2014}                       & + &   & + & + &   &   & + &   &    & + &   &   &   &   &   & \hskip 1sp \cite{code_MPM}       &                                          \\
		\Salg{pm}~\cite{zass_probabilistic_2008}                 & + &   & + & + & + &   &   & + &    & + &   &   &   &   &   & \hskip 1sp \cite{code_FGMD}      &                                          \\
		\Salg{rrwm}~\cite{RandomWalksForGraphMatching}           & + &   & + &   &   & + &   & + &    & + &   &   &   &   &   & \hskip 1sp \cite{caltech}        &                                          \\
		\Salg{smac}~\cite{cour2007balanced}                      & + &   & + & + &   &   & + &   & +  & + &   &   &   &   &   & \hskip 1sp \cite{code_cour}      &                                          \\
		\Salg{sm}~\cite{leordeanu_spectral_2005}                 & + &   & + & + &   &   & + &   & +  & + &   &   &   &   &   & \hskip 1sp \cite{code_leordeanu} &                                          \\
		\midrule
		\Salg{dd-ls(0/3/4)}~\cite{GraphMatchingDDTorresaniEtAl}  &   & + &   &   &   &   &   &   &    &   &   &   & + & + &   &                                  & \hskip 1sp \cite{code_kolmogorov}        \\
		\Salg{fm-bca}~\cite{hutschenreiter_fusionmoves_2021}     &   & + &   &   &   &   &   &   &    &   &   & + & + &   & + &                                  & \hskip 1sp \cite{fusionMovesProjectPage} \\
		\Salg{hbp}~\cite{HungarianBP}                            &   & + & + &   &   &   &   &   &    &   &   &   & + &   & + & \hskip 1sp \cite{code_zhang}     &                                          \\
		\Salg{mp(-mcf/-fw)}~\cite{swoboda2017study}              &   & + &   &   &   &   &   &   &    &   &   &   & + &   & + &                                  & \hskip 1sp \cite{code_swoboda}           \\
		\bottomrule
	\end{tabular}%
	\hfill
	\begin{minipage}{60mm}
		\textbf{Meaning of properties}
		(`+' indicates presence):
		\emph{IQP}: addresses IQP formulation;
		\emph{ILP}: addresses ILP formulation;
		\emph{bijective}: addresses bijective formulation;
		\emph{non-pos.}: requires non-positive costs, see~\cref{rem:non-pos};
		\emph{0-unary}: requires zero unary costs;
		\emph{lineariz.}: linearization-based method;
		\emph{norm}: imposes norm-constraints;
		\emph{doubly}: addresses doubly-stochastic relaxation;
		\emph{spectral}: solves spectral relaxation;
		\emph{discret.}: discretization as in~\cref{rem:discretization};
		\emph{path fol.}: path following method;
		\emph{fusion}: utilizes fusion;
		\emph{duality}: Lagrange duality-based;
		\emph{SGA}: uses dual sub-gradient ascent;
		\emph{BCA}: uses dual block-coordinate ascent;
		\emph{Matlab}: implemented in Matlab [reference to code];
		\emph{\Cpp}: implemented in \Cpp [reference to code].
	\end{minipage}
	\caption{
		\textbf{Method properties.}
		Purely primal heuristics are separated from the dual methods by a horizontal line.
		\label{tab:methods}
	}
	\vspace*{-1mm}
\end{table*}

%% file: floats/table-bca-algorithms.tex
\begin{table*}[t]
	\sffamily
	\begin{tabular}{llll}
		\toprule
		& (i)~graphical model & (ii)~LAP & (iii)~primal \\
		\midrule
		\Salg{hbp}
			& MPLP~\cite{MPLP}
			& Hungarian~\cite{kuhn1955hungarian}
			& branch\,\&\,bound \\[2pt]
		\Salg{mp(-mcf)}
			& \rdelim\}{2}{*}[\parbox{30mm}{\RaggedRight\sffamily anisotropic diffusion~\cite{savchynskyy2019discrete}}]
			& \rdelim\}{2}{*}[\parbox{20mm}{\RaggedRight\sffamily network simplex~\cite{ahuja1993network}}] & LAP \\[2pt]
		\Salg{mp-fw}    &  &  & \Salg{fw} \\[2pt]
		\Salg{fm-bca}   & MPLP\Rplus\Rplus~\cite{tourani2018mplp} & BCA & \Salg{fm} \\
		\bottomrule
	\end{tabular}%
	\hfill
	\begin{minipage}{65mm}
		\small
		Algorithms used for optimizing the \emph{graphical model} and \emph{LAP} part, as well as technique used to obtain a \emph{primal} solution.
		For \emph{LAP} in the \emph{primal} column the solution of the LAP subproblem is reused as feasible assignment.
		Instead of solving the LAP subproblem, \Salg{fm-bca} performs a series of \emph{BCA} steps \wrt~the LAP dual variables.
	\end{minipage}
	\caption{
		Characterization of dual BCA algorithms.
		\label{tab:bca-algorithms}
	}
\end{table*}

%% file: floats/table-datasets.tex
\begin{table*}[t]
	\small\sffamily
	\setlength\tabcolsep{2.5pt}%
	\setlength\aboverulesep{0pt}%
	\setlength\belowrulesep{0pt}%
	\begin{tabular}{@{} l *5{gc} gc @{}}
		\toprule
		dataset
			& \rotatebox{90}{\#inst.}
			& \rotatebox{90}{\#opt.}
			& \rotatebox{90}{bijective}
			& \rotatebox{90}{injective}
			& \rotatebox{90}{non-pos.}
			& \rotatebox{90}{0-unary~}
			& \rotatebox{90}{$|\SV|$}
			& \rotatebox{90}{$|\SL|/|\SV|$}
			& \rotatebox{90}{\parbox{15mm}{\centering density\\(\%)}}
			& \rotatebox{90}{\parbox{15mm}{\centering diagonal\\dens.\ (\%)}}
			& \rotatebox{90}{data} \\
		\midrule
		\Sdata{caltech-large}\cite{RandomWalksForGraphMatching}                   &   9 &  1  &   & + & + & + &  36-219 &       0.4-2.3 &  0.55    & 18.1 & \hskip 1sp \cite{caltech}                              \\
		\Sdata{caltech-small}\cite{RandomWalksForGraphMatching}                   &  21 &  12 &   & + & + & + &   9-117 &         0.4-3 &  0.99    & 26.8 & \hskip 1sp \cite{caltech}                              \\
		\Sdata{car}\cite{UnsupervisedLearningForGraphMatching,VocPascal}          &  30 &  30 & + &   & + &   &   19-49 &             1 &  2.9     & 100  & \hskip 1sp \cite{cars_motor}                           \\
		\Sdata{flow}\cite{GraphFlow,swoboda2017study}                             &   6 &  6  &   &   &   &   &  48-126 & $\approx$   1 &  0.39    & 15.8 & \hskip 1sp \cite{graphflow_problems,graphflow_images}  \\
		\Sdata{hotel}\cite{CaetanoMCLS09,GraphMatchingDDTorresaniEtAl}            & 105 & 105 & + &   &   &   &      30 &             1 & 12.8     & 100  & \hskip 1sp \cite{caetano_2011_data,house_hotel_models} \\
		\Sdata{house-dense}\cite{CaetanoMCLS09,GraphMatchingDDTorresaniEtAl}      & 105 & 105 & + &   &   &   &      30 &             1 & 12.6     & 100  & \hskip 1sp \cite{caetano_2011_data,house_hotel_models} \\
		\Sdata{house-sparse}\cite{CaetanoMCLS09,HungarianBP}                      & 105 & 105 & + &   & + & + &      30 &             1 &  1.5     & 100  & \hskip 1sp \cite{caetano_2011_data}                    \\
		\Sdata{motor}\cite{UnsupervisedLearningForGraphMatching,VocPascal}        &  20 &  20 & + &   & + &   &   15-52 &             1 &  3.8     & 100  & \hskip 1sp \cite{cars_motor}                           \\
		\Sdata{opengm}\cite{KomodakisP08,OpenGMBenchmark}                         &   4 &  4  & + &   &   &   &   19-20 &             1 & 74.8     & 100  & \hskip 1sp \cite{opengm}                               \\
		\Sdata{pairs}\cite{kainmueller2014active,hutschenreiter_fusionmoves_2021} &  16 &  0  &   &   &   &   & 511-565 & $\approx$ 1   &  0.0019  &  3.7 & \hskip 1sp \cite{fusionMovesProjectPage}               \\
		\Sdata{worms}\cite{kainmueller2014active}                                 &  30 &  28 &   &   &   &   &     558 & $\approx$ 2.4 &  0.00038 &  1.6 & \hskip 1sp \cite{kainmueller2017graph}                 \\
		\bottomrule
	\end{tabular}%
	\hfill
	\begin{minipage}{55mm}
		\textbf{Meaning of properties:}\\
		\emph{\#inst.}:~number of problem instances;
		\emph{\#opt.}:~number of known optima;
		\emph{bijective/injective}:~bi-/injective assignment is assumed;
		\emph{non-pos.}:~all costs are non-positive;
		\emph{0-unary}:~datasets with zero unary costs;
		$|\SV|$:~number of elements in $\SV$;
		$|\SL|/|\SV|$:~ratio of the number of elements in $\SL$ to the number of elements in $\SV$;
		\emph{density (\%)}:~percentage of non-zero elements in the cost matrix $C$;
		\emph{diag. dens. (\%)}:~percentage of non-infinite elements on the diagonal of C;
		\emph{data}:~[references] to problem instances, images, feature coordinates or ground truth.
	\end{minipage}
	\caption{
		\textbf{Dataset properties.}
		A `+' indicates that all problem instances of the dataset have the respective property.
		\label{tab:datasets}
	}
\end{table*}

%% file: floats/figure-performance-profile.tex
\begin{figure*}[t]
	\sffamily
	\raisebox{-.5\height}{\includegraphics[trim=14 18 20 14, width=103mm]{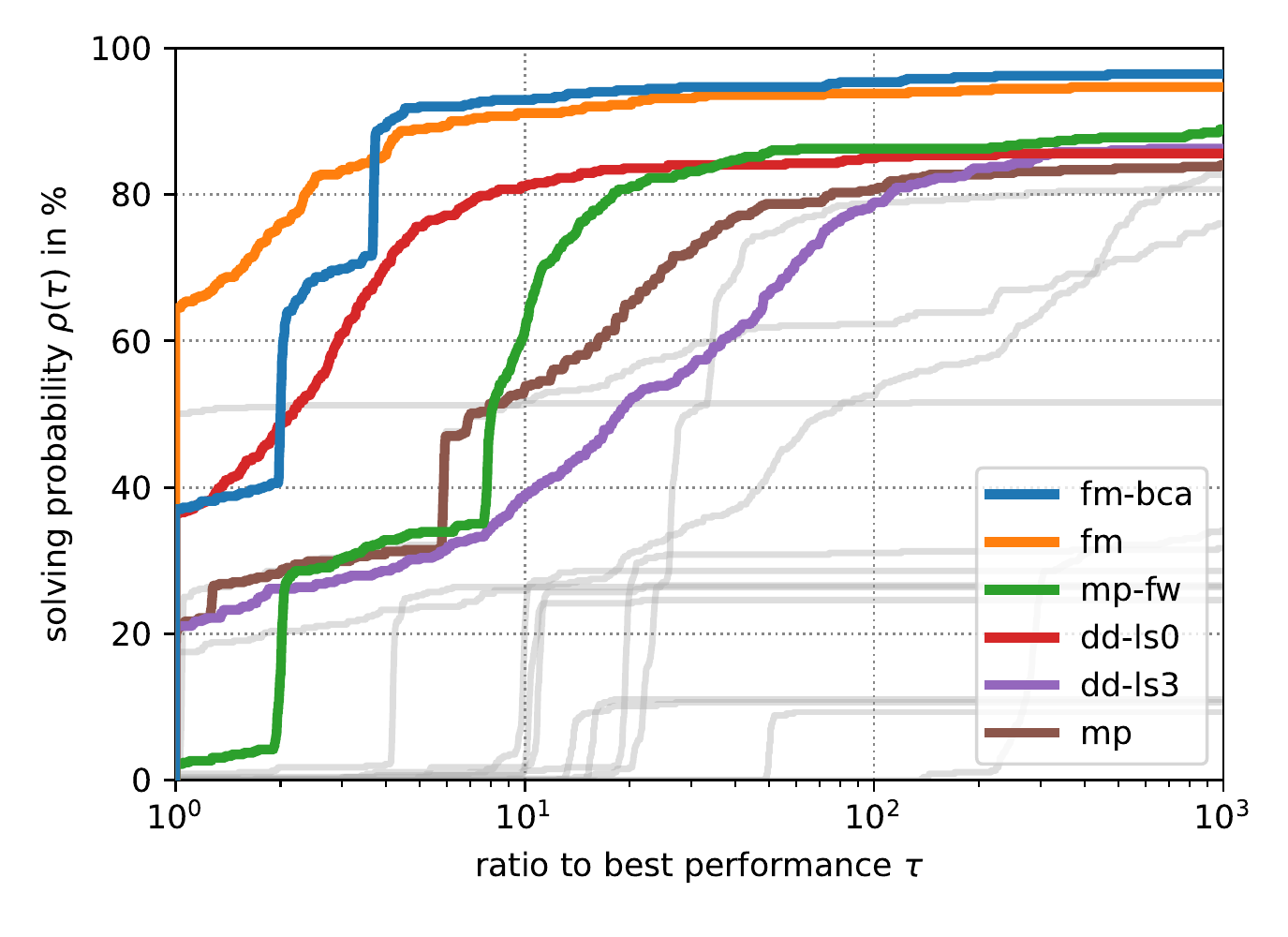}}%
	\hfill
	\begin{minipage}{65mm}
		Methods solving the largest number of instances, see~$\rho(\tau = 10^3)$, are highlighted in color.
		Other methods are shown as ``ghosts'', \ie, unlabeled in gray.
		\Salg{fm} is the best solver in 65\% of all cases, see~$\rho(\tau = 1)$.
		\Salg{fm-bca} outperforms \Salg{fm} when the allowed performance ratio is increased to $\tau$ $\geq$ 3.7.
		Overall, \Salg{fm-bca} solves $\approx$\,97\% and \Salg{fm} solves $\approx$\,95\% of all instances.
		Following are duality-based methods like \Salg{mp-fw} and \Salg{dd-ls0}.

		\vspace{7mm}
	\end{minipage}
	\caption{
		\textbf{Run time performance profile}~\cite{dolan2002benchmarking} across all 451 instances.
		\label{fig:performance-profile}
	}
	\vspace*{-1.5mm}
\end{figure*}

%% file: floats/table-small-problems.tex
\begin{table*}[p]
	\small\sffamily
	\newcommand\MyCols{}
	\setlength{\tabcolsep}{2.9pt}%
	\setlength{\aboverulesep}{0pt}
	\setlength{\belowrulesep}{0pt}
	\def\arraystretch{1.15}
	\centerline{%
	\begin{tabular}{
	        l       
	        *{3}{g} 
	        *{3}{c} 
	        *{3}{g} 
	        *{3}{c} 
	        *{3}{g} 
	        *{2}{c} 
	        *{3}{g} 
	  }
		\toprule
		& \multicolumn{3}{g}{\Sdata{hotel}\,(1s)}
		& \multicolumn{3}{c}{\Sdata{house-dense}\,(1s)}
		& \multicolumn{3}{g}{\Sdata{house-sparse}\,(1s)}
		& \multicolumn{3}{c}{\Sdata{car}\,(1s)}
		& \multicolumn{3}{g}{\Sdata{motor}\,(1s)}
		& \multicolumn{2}{c}{\Sdata{opengm}\,(1s)}
		& \multicolumn{3}{g}{\Sdata{caltech-small}\,(1s)}
		 \\
		& \MyCols{opt} & \MyCols{\textit{E}} & \MyCols{acc} 
		& \MyCols{opt} & \MyCols{\textit{E}} & \MyCols{acc} 
		& \MyCols{opt} & \MyCols{\textit{E}} & \MyCols{acc} 
		& \MyCols{opt} & \MyCols{\textit{E}} & \MyCols{acc} 
		& \MyCols{opt} & \MyCols{\textit{E}} & \MyCols{acc} 
		& \MyCols{opt} & \MyCols{\textit{E}}                
		& \MyCols{opt} & \MyCols{\textit{E}} & \MyCols{acc} \\ 
		\midrule
		\input{generated-tables/small-problems} 
		\bottomrule
	\end{tabular}}

	\smallskip
	\textbf{opt:} optimally solved instances (\%);
	\textbf{E:} average best objective value;
	\textbf{acc:} average accuracy corresponding to best objective (\%)

	\textbf{---*:} method yields no solution for at least one problem instance within the given time interval.

	\caption{
		\textbf{Fixed-time evaluation of small problem instances.}
		Maximal run-time per problem instance is 1 second.
		Boldface marks best values, except for accuracy since algorithm do not optimize it explicitly and do not have access to ground truth.
		Horizontal line separates purely primal from duality-based methods.
		Accuracy omitted for \Sdata{opengm} as no ground truth available.
		Dual bounds omitted as most problems are solved optimally.
		\label{tab:comparison-small-datasets}
	}
\end{table*}

%% file: generated-tables/small-problems.tex
\Salg{fgmd} & 0 & \multicolumn{2}{g}{---*} & 0 & \multicolumn{2}{c}{---*} & 0 & \multicolumn{2}{g}{---*} & 0 & \multicolumn{2}{c}{---*} & 0 & \multicolumn{2}{g}{---*} & 0 & \multicolumn{1}{c}{---*} & 0 & \multicolumn{2}{g}{---*} \\
\Salg{fm} & 97 & -4292 & 100 & \bfseries 100 & \bfseries -3778 & 100 & \bfseries 100 & \bfseries -67 & 100 & 77 & \bfseries -69 & 88 & 90 & \bfseries -63 & 93 & \bfseries 100 & \bfseries -171 & \bfseries 52 & -8906 & 58 \\
\Salg{fw} & 97 & -4288 & 99 & \bfseries 100 & \bfseries -3778 & 100 & 0 & 0 & 0 & 7 & -63 & 63 & 20 & -58 & 70 & 0 & -152 & 0 & 0 & 0 \\
\Salg{ga} & 0 & 947 & 15 & 0 & 3491 & 8 & \bfseries 100 & \bfseries -67 & 100 & 57 & -68 & 84 & 55 & -62 & 90 & 50 & -167 & 0 & \multicolumn{2}{g}{---*} \\
\Salg{ipfps} & 0 & 1051 & 14 & 0 & 3654 & 8 & \bfseries 100 & \bfseries -67 & 100 & 10 & -65 & 80 & 25 & -61 & 85 & 0 & -95 & 19 & \bfseries -8983 & 67 \\
\Salg{ipfpu} & 0 & 1062 & 15 & 0 & 3659 & 8 & \bfseries 100 & \bfseries -67 & 100 & 7 & -60 & 69 & 15 & -58 & 77 & 0 & -86 & 10 & -8829 & 62 \\
\Salg{lsm} & 0 & \multicolumn{2}{g}{---*} & 0 & \multicolumn{2}{c}{---*} & 46 & -65 & 96 & 0 & -51 & 52 & 10 & -52 & 64 & 0 & -67 & 0 & \multicolumn{2}{g}{---*} \\
\Salg{mpm} & 43 & -2585 & 78 & 0 & 1260 & 53 & 0 & -60 & 90 & 7 & \multicolumn{2}{c}{---*} & 5 & \multicolumn{2}{g}{---*} & 0 & -94 & 0 & \multicolumn{2}{g}{---*} \\
\Salg{pm} & 0 & 775 & 33 & 0 & 3262 & 18 & 0 & -54 & 83 & 0 & -35 & 23 & 0 & -35 & 32 & 0 & -83 & 0 & -6510 & 51 \\
\Salg{rrwm} & 0 & 744 & 15 & 0 & 2895 & 10 & \bfseries 100 & \bfseries -67 & 100 & 37 & -68 & 87 & 50 & -62 & 89 & 0 & -154 & 5 & \multicolumn{2}{g}{---*} \\
\Salg{sm} & 0 & 1086 & 13 & 0 & 3789 & 9 & 96 & \bfseries -67 & 100 & 7 & -63 & 76 & 40 & -60 & 87 & 0 & -101 & 0 & -3932 & 36 \\
\Salg{smac} & 1 & -1571 & 61 & 0 & 2817 & 31 & 37 & -44 & 63 & 0 & -52 & 52 & 10 & -52 & 66 & 0 & -84 & 0 & -6196 & 42 \\
\midrule
\Salg{dd-ls0} & \bfseries 100 & \bfseries -4293 & 100 & \bfseries 100 & \bfseries -3778 & 100 & \bfseries 100 & \bfseries -67 & 100 & \bfseries 97 & \bfseries -69 & 91 & \bfseries 100 & \bfseries -63 & 97 & 50 & -160 & 43 & -7414 & 58 \\
\Salg{dd-ls3} & \bfseries 100 & \bfseries -4293 & 100 & \bfseries 100 & \bfseries -3778 & 100 & 96 & \bfseries -66 & 100 & 47 & -57 & 74 & 65 & -57 & 87 & 0 & -118 & 33 & -6842 & 57 \\
\Salg{dd-ls4} & 98 & -4291 & 100 & 92 & -3763 & 99 & 18 & -56 & 86 & 3 & -49 & 59 & 30 & -52 & 78 & 0 & -105 & 24 & -6332 & 54 \\
\Salg{fm-bca} & \bfseries 100 & \bfseries -4293 & 100 & \bfseries 100 & \bfseries -3778 & 100 & \bfseries 100 & \bfseries -67 & 100 & 93 & \bfseries -69 & 92 & \bfseries 100 & \bfseries -63 & 97 & 75 & -170 & 38 & -8927 & 62 \\
\Salg{hbp} & 97 & \multicolumn{2}{g}{---*} & 98 & \multicolumn{2}{c}{---*} & \bfseries 100 & \bfseries -67 & 100 & 77 & \multicolumn{2}{c}{---*} & 95 & \multicolumn{2}{g}{---*} & 0 & \multicolumn{1}{c}{---*} & 0 & \multicolumn{2}{g}{---*} \\
\Salg{mp} & 93 & -4280 & 99 & 99 & -3777 & 100 & \bfseries 100 & \bfseries -67 & 100 & 80 & \bfseries -69 & 92 & 90 & \bfseries -63 & 96 & 0 & -57 & 14 & -7967 & 59 \\
\Salg{mp-fw} & 99 & -4292 & 100 & \bfseries 100 & \bfseries -3778 & 100 & \bfseries 100 & \bfseries -67 & 100 & 90 & \bfseries -69 & 91 & 95 & \bfseries -63 & 98 & 0 & -150 & 33 & -8886 & 60 \\
\Salg{mp-mcf} & 90 & -4245 & 98 & 31 & -3542 & 89 & \bfseries 100 & \bfseries -67 & 100 & 87 & \bfseries -69 & 91 & 90 & \bfseries -63 & 98 & 0 & -57 & 5 & -7882 & 60 \\

%% file: floats/table-large-problems.tex
\begin{table*}[p]
  \small\sffamily
  \setlength{\tabcolsep}{1.8pt}
  \setlength{\aboverulesep}{0pt}
  \setlength{\belowrulesep}{0pt}
  \def\arraystretch{1.15}
  \centerline{%
  \begin{tabular}{l
                  *{3}{g} 
                  *{4}{c} 
                  *{3}{g} 
                  *{3}{c} 
                  *{3}{g} 
                  *{3}{c} 
                 }
    \toprule
    & \multicolumn{3}{g}{\Sdata{flow}\,(1s)}
    & \multicolumn{4}{c}{\Sdata{worms}\,(1s)}
    & \multicolumn{3}{g}{\Sdata{caltech-large}\,(10s)}
    & \multicolumn{3}{c}{\Sdata{caltech-large}\,(100s)}
    & \multicolumn{3}{g}{\Sdata{pairs}\,(10s)}
    & \multicolumn{3}{c}{\Sdata{pairs}\,(100s)}\\
    & opt & \textit{E} & \textit{D} 
    & opt & \textit{E} & \textit{D}  & acc 
    & \textit{E} & \textit{D}  & acc 
    & \textit{E} & \textit{D}  & acc 
    & \textit{E} & \textit{D}  & acc 
    & \textit{E} & \textit{D}  & acc \\ 
    \midrule
    \input{generated-tables/large-problems} 
    \bottomrule
  \end{tabular}}

  \smallskip
  \textbf{opt, E, acc, ---*:} same as in \cref{tab:comparison-small-datasets};\hfill
  \textbf{D:} best attained lower bound if applicable, \ie, for dual methods, otherwise --

  \caption{%
    \textbf{Fixed-time evaluation of mid-size and large problem instances.} Only the best performing algorithms are shown.
    Notation is the same as in Table~\ref{tab:comparison-small-datasets}.
    For each dataset the maximum allowed run-time per instance is given in parentheses.
    For \Sdata{flow} no ground truth is available, so the column \textit{acc} is omitted.
    For \Sdata{caltech-large} and \Sdata{pairs} no global optima are known, and the column \textit{opt} is omitted.
    \label{tab:comparison-large-datasets}
  }
\end{table*}

%% file: generated-tables/large-problems.tex
\Salg{fm} & \bfseries 83 & \bfseries -2838 & -3436 & \bfseries 93 & -48457 & -55757 & 89 & -34117 & -142829 & 52 & -34125 & -142829 & 52 & -65625 & -76418 & 54 & -65825 & -76418 & 55 \\
\Salg{fw} & 67 & -2828 & -- & 0 & -46974 & -- & 81 & 0 & -- & 0 & 0 & -- & 0 & \bfseries -65797 & -- & 54 & -65802 & -- & 54 \\
\midrule
\Salg{dd-ls0} & 33 & -2345 & -2968 & 0 & 60443 & -163870 & 26 & -32973 & \bfseries-35007 & 51 & -33539 & -34959 & 52 & -61482 & -73521 & 41 & -62974 & \bfseries-67306 & 57 \\
\Salg{dd-ls3} & 17 & -2059 & -3030 & 0 & 64017 & -160520 & 24 & -28653 & -42079 & 49 & -33552 & \bfseries-34914 & 49 & -61638 & -73528 & 41 & -62426 & -67599 & 50 \\
\Salg{dd-ls4} & 0 & -2062 & -3090 & 0 & 65731 & -160409 & 24 & -25599 & -62120 & 46 & -30148 & -38880 & 51 & -61634 & -74053 & 41 & -61634 & -70214 & 41 \\
\Salg{fm-bca} & \bfseries 83 & \bfseries -2838 & -2898 & \bfseries 93 & \bfseries -48460 & \bfseries-48514 & 89 & -34040 & -48223 & 51 & -34073 & -48217 & 51 & -65567 & -70163 & 55 & \bfseries -65913 & -69003 & 58 \\
\Salg{mp} & 33 & -2628 & \bfseries-2887 & 0 & \multicolumn{3}{c}{---*} & -32017 & -46070 & 48 & -32069 & -46066 & 48 & -64150 & \bfseries-68255 & 57 & -64380 & -68136 & 57 \\
\Salg{mp-fw} & \bfseries 83 & \multicolumn{2}{g}{---*} & 0 & \multicolumn{3}{c}{---*} & \bfseries -34237 & -48882 & 51 & \bfseries -34277 & -45923 & 51 & \multicolumn{3}{g}{---*} & \multicolumn{3}{c}{---*} \\
\Salg{mp-mcf} & 33 & -2521 & -2892 & 0 & \multicolumn{3}{c}{---*} & -30362 & -46630 & 47 & -30737 & -43833 & 47 & -63990 & -68318 & 56 & -64174 & -68053 & 57 \\

%% file: appendix.tex
\cleardoublepage
\thispagestyle{plain.scrheadings}
\twocolumn[
  \begin{center}
    \leftskip=5mm plus 1fil%
    \rightskip=5mm plus 1fil%
    \bfseries\huge
    Supplementary Material -- A~Comparative Study of Graph Matching Algorithms in Computer~Vision
  \end{center}
  \vspace{2cm}
]

\cohead{Supplementary Material -- A Comparative Study of Graph Matching Algorithms in Computer Vision}

\setcounter{equation}{0}
\setcounter{figure}{0}
\setcounter{remark}{0}
\setcounter{section}{0}
\setcounter{table}{0}
\def\theequation{A\arabic{equation}}
\def\thefigure{A\arabic{figure}}
\def\theremark{A\arabic{remark}}
\def\thesection{A\arabic{section}}
\def\thetable{A\arabic{table}}

\section{Equivalence Proofs}

\begin{theorem}
  \label{supp-thm:equivalence}
  The graph matching problem and the quadratic assignment problem (QAP) are polynomially reducible to each other.
\end{theorem}

While the formal proof can be found below, let us develop an intuition for the problem behind \cref{supp-thm:equivalence}.
In general, the main difference between graph matching and the QAP is that while the QAP requires complete matchings, graph matching allows for incomplete matchings, \ie, not every element of the first set has to be assigned to one of the second and vice versa.
So the equivalence construction mainly has to deal with these side constraints.

When transitioning from graph matching to the QAP, we go from incomplete to complete matchings.
Therefore, the idea is to extend the two sets, usually called $\SV$ and $\SL$, by ``dummy'' elements to which all previously unassigned elements can be assigned without changing the total cost, see \cref{supp-fig:gm-qap} for an illustration.

On the other hand, when going from the QAP to graph matching, we switch from a complete to an incomplete matching.
Here, the idea is to shift the cost structure in such a way that the cost difference of any two complete matchings stays the same, while at the same time guaranteeing that any incomplete matching can be improved by completing it, see \cref{supp-fig:qap-gm} for an illustration.

\begin{proof}
  \allowdisplaybreaks
  We will prove \cref{supp-thm:equivalence} by construction.

  \textbf{Reduction of graph matching to QAP.}
  Consider the graph matching problem, \cf~\eqref{equ:graph-matching-def},
  \begin{align}
      \min_{x\,\in\, \{0,1\}^{\SV\times\SL}} & \sum _{\substack{i,j\,\in\,\SV\\s,l\,\in\,\SL}} c_{is,jl}\, x_{is}\, x_{jl} \label{supp-eq:gm}\\
      \text{subject to}
          & \sum _{s\,\in\,\SL} x_{is} \leq 1 \quad\text{for all $i\in\SV$}, \label{supp-eq:gm-constr}\\
          & \sum _{i\,\in\,\SV} x_{is} \leq 1 \quad\text{for all $s\in\SL$} \nonumber
  \end{align}
  for finite sets $\SV$ and $\SL$, and cost function $c\colon (\SV\times\SL)^2 \to\mathbb{R}$.

  Let $M$ be the disjoint union of $\SV$ and $\SL$, $M:=\SV \dotcup \SL$, and let $w\colon M^4\to\mathbb{R}$ be defined as
  \begin{align*}
      w_{is,jl} = w(i,s,j,l) :=
          \begin{cases}
              c_{is,jl}, & \text{if $i,j\in\SV$, $s,l\in\SL$}, \\
              0, & \text{otherwise}.
          \end{cases}
  \end{align*}
  We can then formulate the QAP
  \begin{align}
      \min_{y\,\in\, \{0,1\}^{M\times M}} & \sum _{\substack{i,j\,\in\, M\\s,l\,\in\, M}} w_{is,jl}\, y_{is}\, y_{jl} \label{supp-eq:qap}\\
      \text{subject to}
          & \sum _{s\,\in\, M} y_{is} = 1 \quad\text{for all $i\in M$}, \label{supp-eq:qap-constr}\\
          & \sum _{i\,\in\, M} y_{is} = 1 \quad\text{for all $s\in M$}. \nonumber
  \end{align}
  Let $y^{\star}$ be a solution of \eqref{supp-eq:qap} satisfying \eqref{supp-eq:qap-constr}, \ie,
  \begin{align*}
      y^{\star} \,\in\, \argmin_{y\,\in\, \{0,1\}^{M\times M}} & \sum _{\substack{i,j\,\in\, M\\s,l\,\in\, M}} w_{is,jl}\, y_{is}\, y_{jl} \\
      \text{s.\,t.}\quad
          & \sum _{s\,\in\, M} y_{is} = 1 \quad\text{for all $i\in M$}, \\
          & \sum _{i\,\in\, M} y_{is} = 1 \quad\text{for all $s\in M$}.
  \end{align*}
  Now let $x^{\star}\in\{0,1\}^{\SV\times\SL}$ be defined as $x_{is}^{\star} = y_{is}^{\star}$ for all $i\in\SV$, $s\in\SL$. It remains to show that $x^{\star}$ then is a solution of \eqref{supp-eq:gm} satisfying \eqref{supp-eq:gm-constr}.

\input{floats/appendix-figure-qap}
\input{floats/appendix-figure-gm}

  First note that $x^{\star}$ satisfies the constraints \eqref{supp-eq:gm-constr} since for all $i\,\in\,\SV$
  \begin{align*}
      \sum_{s\,\in\,\SL} x_{is}^{\star}
          &= \sum_{s\,\in\,\SL} y_{is}^{\star} \\
          &\leq \sum_{s\,\in\, M} y_{is}^{\star} && \text{as $\SL\subseteq M$ and $y^{\star}\in\{0,1\}^{M\times M}$}, \\
          &\leq 1 && \text{as $y^{\star}$ satisfies \eqref{supp-eq:qap-constr}}.
  \end{align*}
  Analogously, $\sum_{i\,\in\,\SV} x_{is}^{\star} \leq 1$ for all $s\in\SL$.

  Suppose now that $x^{\star}$ is not a solution of \eqref{supp-eq:gm}, \ie, there exists $x'\in\{0,1\}^{\SV\times\SL}$ satisfying \eqref{supp-eq:gm-constr} with
  \begin{align*}
      \sum _{\substack{i,j\,\in\,\SV\\s,l\,\in\,\SL}} c_{is,jl}\, x'_{is}\, x'_{jl} < \sum _{\substack{i,j\,\in\,\SV\\s,l\,\in\,\SL}} c_{is,jl}\, x^{\star}_{is}\, x^{\star}_{jl}.
  \end{align*}
  We can then define $y'\in\{0,1\}^{M\times M}$ by
  \begin{align*}
      y'_{is} :=
          \begin{cases}
              x'_{is}, & \text{if $i\in\SV$, $s\in\SL$}, \\
              x'_{si}, & \text{if $i\in\SL$, $s\in\SV$}, \\
              1, & \text{if $i\in\SV$, $s=i$, and $\sum_{l\in\SL} x'_{il} < 1$}, \\
              1, & \text{if $i\in\SL$, $s=i$, and $\sum_{j\in\SV} x'_{js} < 1$}, \\
              0, & \text{otherwise}.
          \end{cases}
  \end{align*}
  Observe that $y'$ satisfies the constraints \eqref{supp-eq:qap-constr} since for all $i\in M$
  \begin{align*}
      \sum_{s\,\in\, M} y'_{is}
          &= \sum_{s\,\in\,\SV} y'_{is} +  \sum_{s\,\in\,\SL} y'_{is} \\
          &= \begin{cases}
              y'_{ii} +  \sum_{s\,\in\,\SL} x'_{is} & \text{if $i\in\SV$} \\
              \sum_{s\,\in\,\SV} x'_{si} +  y'_{ii} & \text{if $i\in\SL$} \\
          \end{cases} \\
          &= \begin{cases}
              1 - \sum_{l\,\in\,\SL} x'_{il} +  \sum_{s\,\in\,\SL} x'_{is} & \text{if $i\in\SV$} \\
              \sum_{s\,\in\,\SV} x'_{si} +  1 - \sum_{j\,\in\,\SV} x'_{ji} & \text{if $i\in\SL$} \\
          \end{cases} \\
          &= 1,
  \end{align*}
  and, analogously, $\sum_{i\,\in\, M} y'_{is} = 1$ for all $s\in M$. Then we obtain
  \begin{align*}
    \MoveEqLeft \sum_{\substack{i,j\,\in\, M\\s,l\,\in\, M}} w_{is,jl}\, y'_{is}\, y'_{jl} \\
          &= \sum _{\substack{i,j\,\in\,\SV\\s,l\,\in\,\SL}} c_{is,jl}\, y'_{is}\, y'_{jl} && \text{by definition of $w$}, \\
          &= \sum _{\substack{i,j\,\in\,\SV\\s,l\,\in\,\SL}} c_{is,jl}\, x'_{is}\, x'_{jl} && \text{by definition of $y'$}, \\
          &< \sum _{\substack{i,j\,\in\,\SV\\s,l\,\in\,\SL}} c_{is,jl}\, x^{\star}_{is}\, x^{\star}_{jl} && \text{due to the choice of $x'$}, \\
          &= \sum _{\substack{i,j\,\in\,\SV\\s,l\,\in\,\SL}} c_{is,jl}\, y^{\star}_{is}\, y^{\star}_{jl} && \text{by definition of $x^{\star}$}, \\
          &= \sum _{\substack{i,j\,\in\, M\\s,l\,\in\, M}} w_{is,jl}\, y^{\star}_{is}\, y^{\star}_{jl} && \text{by definition of $w$}, \\
          &\leq \sum _{\substack{i,j\,\in\, M\\s,l\,\in\, M}} w_{is,jl}\, y'_{is}\, y'_{jl} && \text{since $y^{\star}$ solution of \eqref{supp-eq:qap}},
  \end{align*}
  which is a contradiction. Hence, $x^{\star}$ is a solution of \eqref{supp-eq:gm}.

  Therefore, graph matching is polynomially reducible to the quadratic assignment problem as the size of the constructed QAP is polynomial in the size of the initial graph matching problem, and each solution of the QAP directly induces a solution to the graph matching problem.

  \textbf{Reduction of QAP to graph matching.}
  Consider the QAP
  \begin{align}
      \min_{x\,\in\, \{0,1\}^{M\times M}} & \sum _{\substack{i,j\,\in\, M\\s,l\,\in\, M}} w_{is,jl}\, x_{is}\, x_{jl} \label{supp-eq:qap2}\\
      \text{subject to}
          & \sum _{s\,\in\, M} x_{is} = 1 \quad\text{for all $i\in M$}, \label{supp-eq:qap2-constr}\\
          & \sum _{i\,\in\, M} x_{is} = 1 \quad\text{for all $s\in M$}, \nonumber
  \end{align}
  for a finite set $M$ and cost function $w\colon M^4\to\mathbb{R}$.

  We can now formulate the graph matching problem
  \begin{align}
      \min_{x\,\in\, \{0,1\}^{ M\times M}} & \sum _{\substack{i,j\,\in\, M\\s,l\,\in\, M}} c_{is,jl}\, x_{is}\, x_{jl} \label{supp-eq:gm2}\\
      \text{subject to}
          & \sum _{s\,\in\, M} x_{is} \leq 1 \quad\text{for all $i\in M$}, \label{supp-eq:gm2-constr}\\
          & \sum _{i\,\in\, M} x_{is} \leq 1 \quad\text{for all $s\in M$}, \nonumber
  \end{align}
  where $\SV = \SL = M$, and $c_{is,jl} := w_{is,jl} - \max (w) - 1$ for all $i,j,s,l \in M$. Note that by definition $c_{is,jl} < 0$ for all $i,j,s,l \in M$.

  Let $x^{\star}$ be a solution of \eqref{supp-eq:gm2} satisfying \eqref{supp-eq:gm2-constr}. We now want to prove that $x^{\star}$ also satisfies \eqref{supp-eq:qap2-constr} and is a solution of \eqref{supp-eq:qap2}.

  Suppose that $x^{\star}$ does not satisfy \eqref{supp-eq:qap2-constr}, \ie, there exists $i'\in M$ with $\sum_{s\,\in\, M} x^{\star}_{i's} < 1$. Then there also exists $s'\in M$ with $\sum_{i\,\in\, M} x^{\star}_{is'} < 1$. Observe that $x^{\star}_{i's} = 0$ for all $s\in M$, and $x^{\star}_{is'} = 0$ for all $i\in M$ since $x^{\star}\in\{0,1\}^{M\times M}$.

  We can now define $x'$ as
  \begin{align*}
      x'_{is} = \begin{cases}
          1, & \text{if $i=i'$, $s=s'$}, \\
          x^{\star}_{is}, & \text{otherwise}.
      \end{cases}
  \end{align*}
  Due to the choice of $i'$ and $s'$, $x'$ also satisfies the constraints \eqref{supp-eq:gm2-constr}. Furthermore,
  \begin{align*}
      \sum _{\substack{i,j\,\in\, M\\s,l\,\in\, M}} & c_{is,jl}\, x'_{is}\, x'_{jl} \\
          &= \sum _{\substack{i,j\,\in\, M\\s,l\,\in\, M}} c_{is,jl}\, x^{\star}_{is}\, x^{\star}_{jl}
              + \sum _{\substack{i,s\,\in\, M\\is\neq i's'}} c_{is,i's'}\, x^{\star}_{is} \\
              &\qquad + \sum _{\substack{j,l\,\in\, M\\jl\neq i's'}} c_{i's',jl}\, x^{\star}_{jl}
              + c_{i's',i's'} \\
          &< \sum _{\substack{i,j\,\in\, M\\s,l\,\in\, M}} c_{is,jl}\, x^{\star}_{is}\, x^{\star}_{jl},
  \end{align*}
  since $c_{is,jl} < 0$ for all $i,j,s,l\in M$.
  This contradicts $x^{\star}$ being a solution of \eqref{supp-eq:gm2}. Hence, $x^{\star}$ satisfies \eqref{supp-eq:qap2-constr}.

  Suppose now that $x^{\star}$ is not a solution of \eqref{supp-eq:qap2}, \ie, there exists $\overline{x}\in\{0,1\}^{M\times M}$ satisfying \eqref{supp-eq:qap2-constr} with
  \begin{align*}
      \sum _{\substack{i,j\,\in\, M\\s,l\,\in\, M}} w_{is,jl}\, \overline{x}_{is}\, \overline{x}_{jl} < \sum _{\substack{i,j\,\in\, M\\s,l\,\in\, M}} w_{is,jl}\, x^{\star}_{is}\, x^{\star}_{jl}.
  \end{align*}
  Note that $\overline{x}$ also satisfies \eqref{supp-eq:gm2-constr} as it satisfies \eqref{supp-eq:qap2-constr}.

  Due to the constraints \eqref{supp-eq:qap2-constr}, $x^{\star}$ and $\overline{x}$ have exactly $|M|$ nonzero entries, hence when summing over all $i,j,s,l\in M$, both, $x^{\star}_{is}x^{\star}_{jl}$ and $\overline{x}_{is}\overline{x}_{jl}$, are nonzero exactly $|M|^2$ times. Therefore, we obtain
  \begin{align*}
      \sum _{\substack{i,j\,\in\, M\\s,l\,\in\, M}} &w_{is,jl}\, \overline{x}_{is}\, \overline{x}_{jl} - |M|^2(\max(w)+1) \\
          &< \sum _{\substack{i,j\,\in\, M\\s,l\,\in\, M}} w_{is,jl}\, x^{\star}_{is}\, x^{\star}_{jl} - |M|^2(\max(w)+1) \\
      \sum _{\substack{i,j\,\in\, M\\s,l\,\in\, M}} &\bigr( \,w_{is,jl} - \max(w) - 1\,\bigl)\, \overline{x}_{is}\, \overline{x}_{jl} \\
          &< \sum _{\substack{i,j\,\in\, M\\s,l\,\in\, M}} \bigr( \,w_{is,jl} - \max(w) - 1\,\bigl)\, x^{\star}_{is}\, x^{\star}_{jl} \\
      \sum _{\substack{i,j\,\in\, M\\s,l\,\in\, M}} &c_{is,jl}\, \overline{x}_{is}\, \overline{x}_{jl}
          < \sum _{\substack{i,j\,\in\, M\\s,l\,\in\, M}} c_{is,jl}\, x^{\star}_{is}\, x^{\star}_{jl}
  \end{align*}
  This would contradict $x^{\star}$ being a solution of \eqref{supp-eq:gm2}. Thus, $x^{\star}$ is a solution of \eqref{supp-eq:qap2}.

  This proves that the quadratic assignment problem is polynomially reducible to graph matching as the size of the constructed graph matching problem is polynomial in the size of the initial QAP. Each solution of the graph matching problem directly yields a solution of the QAP.
\end{proof}

\input{floats/appendix-table-dd}

\begin{remark}
Note that a polynomial reduction from the QAP to graph matching can also be obtained by shifting only the unary costs by a sufficiently large constant, \ie, by defining
\begin{align*}
    c_{is,jl} = \begin{cases}
            w_{is,jl} - K, & \text{if $jl=is$, $i,j,s,l\in M$} \\
            w_{is,jl}, & \text{otherwise.} \\
        \end{cases}
\end{align*}
for sufficiently large $K$.
This is relevant in practice as it only changes the diagonal of the cost matrix, and not the full matrix. In this way it does not as heavily influence the sparsity of the cost structure.
\end{remark}

The statements in the proof of \cref{supp-thm:equivalence} hold in particular if all quadratic terms incur a cost of zero, so we can reduce the objective function to its linear component, \ie, the objective function
\begin{align*}
    \sum _{\substack{i,j\,\in\,\SV\\s,l\,\in\,\SL}} c_{is,jl}\, x_{is}\, x_{jl}
\end{align*}
of the graph matching problem, \cf~\eqref{supp-eq:gm} in the proof above, can be reduced to
\begin{align*}
    \sum _{\substack{i\,\in\,\SV\\s\,\in\,\SL}} c_{is}\, x_{is},
\end{align*}
which corresponds to the incomplete linear assignment problem. Similarly, we can replace the objective function
\begin{align*}
    \sum _{\substack{i,j\,\in\, M\\s,l\,\in\, M}} w_{is,jl}\, y_{is}\, y_{jl}
\end{align*}
of the quadratic assignment problem, \cf~\eqref{supp-eq:qap2}, by
\begin{align*}
    \sum _{i,s\,\in\, M} w_{is}\, y_{is},
\end{align*}
which corresponds to the objective of the linear assignment problem. With this in mind, the corollary below directly follows from \cref{supp-thm:equivalence}:

\begin{corollary}\label{supp-cor:equivalence}
    The incomplete linear assignment problem and the linear assignment problem (LAP) are polynomially reducible to each other.
\end{corollary}

\section{Symmetry of the Formulations}

Note that formulations~\eqref{equ:map-formulation} and~\eqref{equ:ILP-representation}-\eqref{equ:coupling-constraints-ILP} are asymmetric, since the sets $\SV$ and $\SL$ play different roles in the formulations, in contrast to the symmetric formulation~\eqref{equ:graph-matching-def}.
Although swapping the roles of $\SV$ and $\SL$ does not influence the optimal solutions, it influences the problem structure, such as the sparsity of the resulting graphical model, as well as the tightness of the corresponding LP relaxation.
Symmetrized formulations based on the graphical model representation use two graphical models, with the role of $\SV$ and $\SL$ being swapped in one of them~\cite{swoboda2017study}.
These graphical models, with variables denoted correspondingly by $x$ and $\hat x$, are coupled by additional constraints which enforce either the unary variables to be equal in both representations, \ie, $x_{is}=\hat x_{si}$, or the pairwise variables, \ie, $x_{is,jl}=\hat x_{si,lj}$.
The first type of coupling constraints is computationally cheaper, whereas the second one leads to a tighter LP relaxation equivalent to the standard one considered in operations research~\cite{adams1994improved}.
In our experimental evaluation the second type of constraints is implicitly used by \Salg{dd-ls*} algorithms.

\section{Data Format (dd-format)}

As a unified data format for all datasets we use the one introduced by~\cite{GraphMatchingDDTorresaniEtAl}.
It encodes a sparse representation of the graph matching problem~\eqref{equ:graph-matching-def} in plain text.
Due to its origin we refer to it as \emph{dd-} or \emph{dual decomposition} data format.
It is used by \Salg{dd-ls*}, as well as \Salg{fm}, \Salg{fm-bca} and \Salg{mp-*} algorithms as a native input.
\Cref{fig:dd} presents an overview of the file format and shows a description of of all input line types.

With other algorithms suitable converters are used.
The one for the graphical model representation~\eqref{equ:map-formulation} is readily available at~\cite{fusionMovesProjectPage}.
For Matlab algorithms a matrix representation of~\eqref{equ:graph-matching-def} is required.
We implemented this conversion by ourselves and will make the code available.

\section{Cost Transformations}

Different methods have different requirements for the cost matrix, see columns \emph{bijective}, \emph{non-positive} and \emph{zero unary} in \cref{tab:methods}.
For datasets where the costs do not fulfill those requirements, see equivalent columns in \cref{tab:datasets}, the costs have to be transformed to make them amenable to those algorithms.
The transformations are chosen to in a way such that
\begin{Itemize}
\Item the cost difference between any two matchings stays the same; and
\Item the sparsity of the cost matrix is preserved as much as possible.
\end{Itemize}
In the following we describe each of the three transformations.
The full implementation is part of our published source code for the benchmark.
Without loss of generality we assume the cost matrix to be upper triangular, \ie $c_{is,jl} = 0$ for $i, j \in \SV$, $s, l \in \SL$ and $i > j$.

\Paragraph{Bijective Costs.}
Non-bijective problem instances are transformed into the bijective ones in a way similar to the described in the proof of Theorem~\ref{supp-thm:equivalence} and illustrated in Fig.~\ref{supp-fig:gm-qap}.

\Paragraph{Non-positive Costs.}
Here we transform the cost matrix into a form where all \emph{finite} costs are non-positive.
Note that infinite costs will stay positive, as they are identified by corresponding algorithms and prohibit selection of the associated matchings.
\Cref{alg:appendix-non-positive} shows the transformation which shifts the costs of all bijective assignments by a fixed constant.
Therefore, for non-bijective instances we first transform them into the bijective form as described in previous paragraph.

\input{floats/appendix-algorithm-non-positive}

\Paragraph{Remove Unary Costs.}
The \Salg{pm} algorithm does not take into account unary costs, see column \emph{zero unary} in \cref{tab:methods}.
We use \cref{alg:zero-unary} to transform unary costs into pairwise costs.

\input{floats/appendix-algorithm-zero-unary}

\ParagraphRun{For the maximization algorithms}
\Salg{ipfpu}, \Salg{ipfps}, \Salg{ga}, \Salg{rrwm}, \Salg{pm}, \Salg{sm}, \Salg{smac}, \Salg{lsm}, \Salg{mpm}, \Salg{fgmd} and \Salg{hbp} the sign of the elements in the cost matrix is flipped additionally after performing all other operations.

\input{floats/appendix-table-cmdline}

\section{Accuracy computation}

For \Sdata{hotel}, \Sdata{house-sparse}, \Sdata{house-dense}, \Sdata{car} and \Sdata{motor} the complete ground truth assignments are known.
The accuracy is computed as the number of correctly assigned nodes over $n_\SV$.
The ground truth for \Sdata{caltech-small}, \Sdata{caltech-large}, \Sdata{worms} and \Sdata{pairs} is only partial i.e. not every node has a ground truth label.
This is taken into account by computing the accuracy as the number of correctly assigned nodes over the number of nodes with available ground truth label.

\section{Run Time Measurement}

To make the comparison as fair as possible, we exclude preprocessing steps and the time it takes to load the graph matching files into the solver. This means that the following steps are excluded from our run-time measurements:
\begin{Itemize}
  \Item Conversion process from the dd file format into the input format of the solvers;
  \Item Loading the file from storage into memory;
  \Item Parsing the input file; and
  \Item Dataset transformation if needed.
\end{Itemize}

Practically, we start the timer directly before the solver starts the optimization routine.
For optimization methods that have not included this functionality, we have modified the methods accordingly.

\section{Reproducibility}

\Paragraph{Original Source Code.}
References with links to the original source code can be found in column \emph{Matlab} and \emph{\Cpp} in \cref{tab:methods}.
Note that for the \Salg{dd-ls*} methods we use the same command line wrapper as used in~\cite{hutschenreiter_fusionmoves_2021}. The wrapper allows to select the problem decomposition at run-time without need for recompilation.
The specific command lines for the \Cpp~programs~(see column \emph{\Cpp} in~\cref{tab:methods}) that have been used in the benchmark are listed in~\cref{tab:cmdline}.

\Paragraph{Matlab Wrappers.}
We incorporated the implementations of the Matlab algorithms in our own wrappers in order to be able to load the datasets and to make the output standardized.
No parameters have to be specified.

\Paragraph{Reproducible Containers.}
For all methods we build Linux containers so that the environment as well as the specific version of the solvers are reproducible.
The container files contain instructions for downloading, processing and building a fixed version  of the corresponding methods.
The source code of each method is fetched from the Internet.
For the case that source code is no longer online, we preserve a historical copy of the repositories.
The container build scripts will pin each method to a specific version or commit id to preserve reproducibility.

\Paragraph{HPC Scripts.}
For usage in high-performance compute clusters~(HPC clusters) we provide a small script to transform the container images into singluarity image.
Most HPC clusters provide tools to easily run these singularity images.
Additionally, we provide SLURM scheduling scripts so that the benchmark can be reproduced easily and quickly.

\Paragraph{Project Web Site.}
The wrappers, container files and other scripts are publicly available on the project web site, see \url{https://vislearn.github.io/gmbench/}.

\section{Detailed Evaluation Results}

More detailed evaluation results are provided on the following pages.
\Cref{fig:appendix-performance-profile} shows run-time performance profiles similar to \cref{fig:performance-profile} but for each dataset separately.
\Crefrange{tab:appendix-full-dataset-caltech-small}{tab:appendix-full-dataset-worms}
show the average results of each method on each dataset for time limits 1s, 10s, 100s and 300s similar to \cref{tab:comparison-small-datasets} and \cref{tab:comparison-large-datasets}.

\input{floats/appendix-figure-performance-profile}

\def\Dataset{caltech-small}
\let\ifHasOpt\iftrue
\let\ifHasAcc\iftrue
\input{floats/appendix-table-full-dataset}

\def\Dataset{caltech-large}
\let\ifHasOpt\iffalse
\let\ifHasAcc\iftrue

\input{floats/appendix-table-full-dataset}
\def\Dataset{car}
\let\ifHasOpt\iftrue
\let\ifHasAcc\iftrue

\input{floats/appendix-table-full-dataset}
\def\Dataset{flow}
\let\ifHasOpt\iftrue
\let\ifHasAcc\iffalse

\input{floats/appendix-table-full-dataset}
\def\Dataset{hotel}
\let\ifHasOpt\iftrue
\let\ifHasAcc\iftrue

\input{floats/appendix-table-full-dataset}
\def\Dataset{house-dense}
\let\ifHasOpt\iftrue
\let\ifHasAcc\iftrue

\input{floats/appendix-table-full-dataset}
\def\Dataset{house-sparse}
\let\ifHasOpt\iftrue
\let\ifHasAcc\iftrue

\input{floats/appendix-table-full-dataset}
\def\Dataset{motor}
\let\ifHasOpt\iftrue
\let\ifHasAcc\iftrue

\input{floats/appendix-table-full-dataset}
\def\Dataset{opengm}
\let\ifHasOpt\iftrue
\let\ifHasAcc\iffalse

\input{floats/appendix-table-full-dataset}
\def\Dataset{pairs}
\let\ifHasOpt\iffalse
\let\ifHasAcc\iftrue

\input{floats/appendix-table-full-dataset}
\def\Dataset{worms}
\let\ifHasOpt\iftrue
\let\ifHasAcc\iftrue

\input{floats/appendix-table-full-dataset}

%% file: floats/appendix-figure-qap.tex
\begin{figure*}[t]
  \centering
  \begin{tikzpicture}
      \foreach \x in {0,1,...,3}
          \node (V\x) [draw, circle, minimum size=1em, fill=LimeGreen, ]
              at (\x, 0) {};
      \foreach \y in {0,1,...,5} {
          \node (L\y) [draw, circle, minimum size=1em, fill=Goldenrod, ]
              at (\y, 3) {};
          \foreach \x in {0,1,...,3}
              \draw []
                  (L\y) -- (V\x);
      }
      \foreach \x in {4,5,...,9} {
          \node (V\x) [draw, circle, minimum size=1em, fill=Goldenrod!30, ]
              at (\x, 0) {};
          \foreach \y in {0,1,...,5}
              \draw [ultra thin, opacity=0.2, ]
                  (L\y) -- (V\x);
      }
      \foreach \y in {6,...,9} {
          \node (L\y) [draw, circle, minimum size=1em, fill=LimeGreen!30, ]
              at (\y, 3) {};
          \foreach \x in {0,1,...,9}
              \draw [ultra thin, opacity=0.2, ]
                  (L\y) -- (V\x);
      }
      \node [fit=(V0) (V3),
          label={[name=labV]left:$\SV$},
          draw=LimeGreen,
          rounded corners=1em,
          inner sep=0.5em,
          line width=3pt,
          draw opacity=0.5,
          ] {};
      \node [fit=(L0) (L5),
          label={[name=labL]left:$\SL$},
          draw=Goldenrod,
          rounded corners=1em,
          inner sep=0.5em,
          line width=3pt,
          draw opacity=0.5,
          ] {};
      \node [fit=(V0) (V9) (labV),
          label={right:$M=\SV \dotcup \SL$},
          draw=WildStrawberry,
          rounded corners=1.5em,
          inner sep=0.5em,
          minimum height=3em,
          line width=1pt,
          ] {};
      \node [fit=(L0) (L9) (labL),
          label={[name=labM]right:$M=\SV \dotcup \SL$},
          draw=WildStrawberry,
          rounded corners=1.5em,
          inner sep=0.5em,
          minimum height=3em,
          line width=1pt,
          draw opacity=0.8,
          ] {};
      \draw [line width=2pt,
          Emerald,
          ]
          ($(V3)+(0.5em,-2em)$) [rounded corners=4em]
          -- ($(L6)+(0.5em,2em)$) [rounded corners=2em]
          -- ($(labL)+(-1.7em,2em)$)
          -- ($(labV)+(-1.7em,-2em)$) [rounded corners=1em]
          -- cycle;
      \draw [line width=2pt,
          rounded corners=2.5em,
          Plum,
          ]
          ($(labM)+(3em,2.5em)$) rectangle ($(labV)+(-2.2em,-2.5em)$);
      \coordinate (ta) at
          ($0.5*(labM)+0.5*(labV)+(0.4em,-2.7)$);
      \node (gm) at ($(ta)+(-2,0)$)
          [anchor=north east,
          ]
          {\begin{minipage}{4.5cm}
              \centering
              \textcolor{Emerald}{\textbf{graph matching problem}}
              \scriptsize\begin{align*}
                  \min_{x\,\in\, \{0,1\}^{\SV\times\SL}} & \sum _{\substack{i,j\,\in\,\SV\\s,l\,\in\,\SL}} c_{is,jl}\, x_{is}\, x_{jl} \\
                  \text{s.\,t.}\quad
                      & \forall\, i\in\SV\colon\;\sum _{s\,\in\,\SL} x_{is} \leq 1 \\ 
                      & \forall\, s\in\SL\colon\;\sum _{i\,\in\,\SV} x_{is} \leq 1 
              \end{align*}
              \normalsize\textcolor{Emerald}{\textbf{solution \textit{x}$^{\star}$}}
          \end{minipage}};
      \node (qap) at ($(ta)+(2,0)$)
          [anchor=north west,
          ]
          {\begin{minipage}{4.5cm}
              \centering
              \textcolor{Plum}{\textbf{constructed QAP}}
              \scriptsize\begin{align*}
                  \min_{y\,\in\, \{0,1\}^{M\times M}} & \sum _{\substack{i,j\,\in\, M\\s,l\,\in\, M}} w_{is,jl}\, y_{is}\, y_{jl} \\
                  \text{s.\,t.}\quad
                      & \forall\,i\in M\colon\;\sum _{s\,\in\, M} y_{is} = 1 \\ 
                      & \forall\, s\in M\colon\;\sum _{i\,\in\, M} y_{is} = 1 
              \end{align*}
              \normalsize\textcolor{Plum}{\textbf{solution \textit{y}$^{\star}$}}
          \end{minipage}};
      \draw [line width=1pt, -latex]
          ($(gm.north east)+(-0.25,-0.75em)$) --
          node [above] {\tiny $M:=\SV \dotcup \SL$}
          node [below] {\tiny $w_{is,jl} :=
                        \begin{cases}
                          c_{is,jl}, & \text{if $i$,$j$$\in$$\SV$, $s$,$l$$\in$$\SL$} \\
                          0, & \text{otherwise}
                        \end{cases}$}
          ($(qap.north west)+(0.25,-0.75em)$);
      \draw [line width=1pt, latex-]
          ($(gm.south east)+(-0.25,0.75em)$) --
          node [above] {\tiny $x^{\star}_{is}:=y^{\star}_{is}$}
          ($(qap.south west|-gm.south east)+(0.25,0.75em)$);
      \fill [Emerald, opacity=0.1]
          ($(gm.north)+(-2.3,-0.6)$) rectangle ++(4.6,-2.9);
      \fill [Plum, opacity=0.1]
          ($(qap.north)+(-2.3,-0.6)$) rectangle ++(4.6,-2.9);
  \end{tikzpicture}

  \caption{
    \textbf{From graph matching to the QAP.}
    The notation corresponds to the notation used in the proof of Theorem \ref{supp-thm:equivalence}.
    The idea behind reducing graph matching to quadratic assignment is to enable completion of the possibly incomplete matchings feasible for graph matching.
    This is done by providing ``dummy'' nodes to which those nodes can be matched that would otherwise be left unassigned.
    As potentially all nodes could be unassigned, we provide a corresponding dummy for each.
    \label{supp-fig:gm-qap}
  }
\end{figure*}

%% file: floats/appendix-figure-gm.tex
\begin{figure*}[t]
  \centering
  \newcommand{\myNN}[1]{%
      \node [blue!#1!white] {};}
  \newcommand{\myPN}[1]{%
      \node [red!#1!white] {};}
  \begin{tikzpicture}
      \coordinate (ta) at (0,0);
      \node (qap) at ($(ta)+(-1.25,0)$)
          [anchor=north east,
          ]
          {\begin{minipage}{4.5cm}
              \centering
              \textcolor{Plum}{\textbf{QAP}}
              \scriptsize\begin{align*}
                  \min_{x\,\in\, \{0,1\}^{M\times M}} & \sum _{\substack{i,j\,\in\, M\\s,l\,\in\, M}} w_{is,jl}\, x_{is}\, x_{jl} \\
                  \text{s.\,t.}\quad
                      & \forall\,i\in M\colon\;\sum _{s\,\in\, M} x_{is} = 1 \\
                      & \forall\,s\in M\colon\;\sum _{i\,\in\, M} x_{is} = 1
              \end{align*}
              \normalsize\textcolor{Plum}{\textbf{solution \textit{x}$^{\star}$}}
          \end{minipage}};
      \node (gm) at ($(ta)+(1.25,0)$)
          [anchor=north west,
          ]
          {\begin{minipage}{4.5cm}
              \centering
              \textcolor{Emerald}{\textbf{constructed GM problem}}
              \scriptsize\begin{align*}
                  \min_{x\,\in\, \{0,1\}^{\SV\times\SL}} & \sum _{\substack{i,j\,\in\,\SV\\s,l\,\in\,\SL}} c_{is,jl}\, x_{is}\, x_{jl} \\
                  \text{s.\,t.}\quad
                      & \forall\,i\in \SV\colon\;\sum _{s\,\in\,\SL} x_{is} \leq 1 \\
                      & \forall\,s\in \SL\colon\;\sum _{i\,\in\,\SV} x_{is} \leq 1
              \end{align*}
              \normalsize\textcolor{Emerald}{\textbf{solution \textit{x}$^{\star}$}}
          \end{minipage}};
      \draw [line width=1pt, -latex]
          ($(qap.north east)+(-0.2,-0.75em)$) --
          node [above] {\tiny $\SV :=M, \SL:=M$}
          node [below=0.1em] {\tiny $c_{is,jl} :=$}
          node [below=0.9em] {\tiny $w_{is,jl}-\max(w)-1$}
          ($(gm.north west)+(0.2,-0.75em)$);
      \draw [line width=1pt, latex-]
          ($(qap.south east)+(-0.2,0.75em)$) --
          ($(gm.south west|-qap.south east)+(0.2,0.75em)$);
      \fill [Emerald, opacity=0.1]
          ($(gm.north)+(-2.35,-0.6)$) rectangle ++(4.7,-2.9);
      \fill [Plum, opacity=0.1]
          ($(qap.north)+(-2.35,-0.6)$) rectangle ++(4.7,-2.9);
      \matrix (w) at ($(qap.north)+(0,0.8)$)
          [nodes={fill, minimum size=1em, anchor=center},
          inner sep=2pt, draw=Plum, line width=2pt,
          anchor=south,]
          {
              \myPN{48} \pgfmatrixnextcell \myNN{4} \pgfmatrixnextcell \myNN{32} \pgfmatrixnextcell \myNN{26} \pgfmatrixnextcell \myPN{41} \pgfmatrixnextcell \myPN{37} \pgfmatrixnextcell \myNN{34} \pgfmatrixnextcell \myPN{40} \pgfmatrixnextcell \myNN{36} \\
              \myPN{39} \pgfmatrixnextcell \myNN{17} \pgfmatrixnextcell \myPN{49} \pgfmatrixnextcell \myNN{37} \pgfmatrixnextcell \myPN{10} \pgfmatrixnextcell \myPN{0} \pgfmatrixnextcell \myNN{24} \pgfmatrixnextcell \myPN{40} \pgfmatrixnextcell \myNN{47} \\
              \myPN{48} \pgfmatrixnextcell \myNN{34} \pgfmatrixnextcell \myNN{42} \pgfmatrixnextcell \myNN{11} \pgfmatrixnextcell \myPN{2} \pgfmatrixnextcell \myPN{22} \pgfmatrixnextcell \myPN{46} \pgfmatrixnextcell \myPN{15} \pgfmatrixnextcell \myPN{13} \\
              \myPN{42} \pgfmatrixnextcell \myNN{48} \pgfmatrixnextcell \myPN{50} \pgfmatrixnextcell \myPN{28} \pgfmatrixnextcell \myPN{31} \pgfmatrixnextcell \myPN{43} \pgfmatrixnextcell \myNN{31} \pgfmatrixnextcell \myPN{32} \pgfmatrixnextcell \myNN{30} \\
              \myNN{30} \pgfmatrixnextcell \myPN{21} \pgfmatrixnextcell \myNN{21} \pgfmatrixnextcell \myPN{12} \pgfmatrixnextcell \myPN{2} \pgfmatrixnextcell \myNN{7} \pgfmatrixnextcell \myNN{19} \pgfmatrixnextcell \myPN{12} \pgfmatrixnextcell \myNN{27} \\
              \myPN{3} \pgfmatrixnextcell \myPN{46} \pgfmatrixnextcell \myNN{24} \pgfmatrixnextcell \myNN{2} \pgfmatrixnextcell \myNN{11} \pgfmatrixnextcell \myNN{49} \pgfmatrixnextcell \myNN{6} \pgfmatrixnextcell \myPN{7} \pgfmatrixnextcell \myNN{6} \\
              \myNN{28} \pgfmatrixnextcell \myPN{6} \pgfmatrixnextcell \myNN{46} \pgfmatrixnextcell \myPN{20} \pgfmatrixnextcell \myPN{28} \pgfmatrixnextcell \myNN{22} \pgfmatrixnextcell \myNN{27} \pgfmatrixnextcell \myPN{43} \pgfmatrixnextcell \myNN{47} \\
              \myNN{21} \pgfmatrixnextcell \myNN{41} \pgfmatrixnextcell \myPN{14} \pgfmatrixnextcell \myPN{44} \pgfmatrixnextcell \myPN{36} \pgfmatrixnextcell \myNN{16} \pgfmatrixnextcell \myNN{42} \pgfmatrixnextcell \myPN{8} \pgfmatrixnextcell \myPN{32} \\
              \myNN{21} \pgfmatrixnextcell \myPN{3} \pgfmatrixnextcell \myNN{35} \pgfmatrixnextcell \myPN{15} \pgfmatrixnextcell \myPN{8} \pgfmatrixnextcell \myNN{48} \pgfmatrixnextcell \myNN{4} \pgfmatrixnextcell \myNN{49} \pgfmatrixnextcell \myNN{3} \\
          };
      \node at (w.south)
          [anchor=north]
          {cost structure $w$};
      \node at (w.west)
          [anchor=south, rotate=90]
          {\scriptsize $is\in M\times M$\vphantom{$j$}};
      \node at (w.north)
          [anchor=south]
          {\scriptsize $jl\in M\times M$};
      \matrix (c) at ($(gm.north)+(0,0.8)$)
          [nodes={fill, minimum size=1em, anchor=center},
          inner sep=2pt, draw=Emerald, line width=2pt,
          anchor=south,]
          {
              \myNN{3} \pgfmatrixnextcell \myNN{55} \pgfmatrixnextcell \myNN{83} \pgfmatrixnextcell \myNN{77} \pgfmatrixnextcell \myNN{10} \pgfmatrixnextcell \myNN{14} \pgfmatrixnextcell \myNN{85} \pgfmatrixnextcell \myNN{11} \pgfmatrixnextcell \myNN{87} \\
              \myNN{12} \pgfmatrixnextcell \myNN{68} \pgfmatrixnextcell \myNN{2} \pgfmatrixnextcell \myNN{88} \pgfmatrixnextcell \myNN{41} \pgfmatrixnextcell \myNN{51} \pgfmatrixnextcell \myNN{75} \pgfmatrixnextcell \myNN{11} \pgfmatrixnextcell \myNN{98} \\
              \myNN{3} \pgfmatrixnextcell \myNN{85} \pgfmatrixnextcell \myNN{93} \pgfmatrixnextcell \myNN{62} \pgfmatrixnextcell \myNN{49} \pgfmatrixnextcell \myNN{29} \pgfmatrixnextcell \myNN{5} \pgfmatrixnextcell \myNN{36} \pgfmatrixnextcell \myNN{38} \\
              \myNN{9} \pgfmatrixnextcell \myNN{99} \pgfmatrixnextcell \myNN{1} \pgfmatrixnextcell \myNN{23} \pgfmatrixnextcell \myNN{20} \pgfmatrixnextcell \myNN{8} \pgfmatrixnextcell \myNN{82} \pgfmatrixnextcell \myNN{19} \pgfmatrixnextcell \myNN{81} \\
              \myNN{81} \pgfmatrixnextcell \myNN{30} \pgfmatrixnextcell \myNN{72} \pgfmatrixnextcell \myNN{39} \pgfmatrixnextcell \myNN{49} \pgfmatrixnextcell \myNN{58} \pgfmatrixnextcell \myNN{70} \pgfmatrixnextcell \myNN{39} \pgfmatrixnextcell \myNN{78} \\
              \myNN{48} \pgfmatrixnextcell \myNN{5} \pgfmatrixnextcell \myNN{75} \pgfmatrixnextcell \myNN{53} \pgfmatrixnextcell \myNN{62} \pgfmatrixnextcell \myNN{100} \pgfmatrixnextcell \myNN{57} \pgfmatrixnextcell \myNN{44} \pgfmatrixnextcell \myNN{57} \\
              \myNN{79} \pgfmatrixnextcell \myNN{45} \pgfmatrixnextcell \myNN{97} \pgfmatrixnextcell \myNN{31} \pgfmatrixnextcell \myNN{23} \pgfmatrixnextcell \myNN{73} \pgfmatrixnextcell \myNN{78} \pgfmatrixnextcell \myNN{8} \pgfmatrixnextcell \myNN{98} \\
              \myNN{72} \pgfmatrixnextcell \myNN{92} \pgfmatrixnextcell \myNN{37} \pgfmatrixnextcell \myNN{7} \pgfmatrixnextcell \myNN{15} \pgfmatrixnextcell \myNN{67} \pgfmatrixnextcell \myNN{93} \pgfmatrixnextcell \myNN{43} \pgfmatrixnextcell \myNN{19} \\
              \myNN{72} \pgfmatrixnextcell \myNN{48} \pgfmatrixnextcell \myNN{86} \pgfmatrixnextcell \myNN{36} \pgfmatrixnextcell \myNN{43} \pgfmatrixnextcell \myNN{99} \pgfmatrixnextcell \myNN{55} \pgfmatrixnextcell \myNN{100} \pgfmatrixnextcell \myNN{54} \\
          };
      \node at (c.south)
          [anchor=north]
          {shifted cost structure $c$};
      \node at (c.west)
          [anchor=south, rotate=90]
          {\scriptsize $is\in\SV\times\SL$\vphantom{$j$}};
      \node at (c.north)
          [anchor=south]
          {\scriptsize $jl\in\SV\times\SL$};
      \draw [top color=red, bottom color=blue, middle color=white,]
          ($0.5*(w.south)+0.5*(c.south)+(-0.125,3pt)$) rectangle ($0.5*(w.north)+0.5*(c.north)+(0.125,-3pt)$);
      \node at ($0.5*(w.south)+0.5*(c.north)+(-0.5,0)$)
          [anchor=center]
          {$0$};
      \node at ($0.5*(w.south)+0.5*(c.north)+(-0.5,1)$)
          [anchor=center]
          {$+$};
      \node at ($0.5*(w.south)+0.5*(c.north)+(-0.5,-1)$)
          [anchor=center]
          {$-$};
  \end{tikzpicture}

  \caption{
    \textbf{From the QAP to graph matching (GM).}
    The notation corresponds to the notation used in the proof of Theorem \ref{supp-thm:equivalence}.
    The idea behind reducing quadratic assignment to graph matching is to make sure that while the requirement for complete matchings is dropped, a better objective value can always be obtained when completing a given incomplete matching.
    This is guaranteed by shifting the cost structure to incur negative costs for all assignments.
    \label{supp-fig:qap-gm}
  }
\end{figure*}

%% file: floats/appendix-table-dd.tex
\begin{table*}[t]
  \sffamily
  \centering
  \begin{tabular}{l l}
    \toprule
    \bfseries input line                   & \bfseries description \\
    \midrule
    \ttfamily c comment line               & comments are ignored\\
    \ttfamily p <\#V> <\#L> <\#A> <\#E>    & prologue: $|\SV|$, $|\SL|$, number of assignments and edges to follow\\
    \ttfamily a <id> <i> <s> <cost>        & specifies possible assignment $i \rightarrow s$ for $i \in \SV$, $s \in \SL$ (unary terms $c_{is,is} x_{is}$)\\
    \ttfamily e <id1> <id2> <cost>         & specifies edge between two assignments (pairwise terms $(c_{is,jl}+c_{jl,is}) x_{is} x_{jl}$)\\
    \ttfamily i0 <i> <x> <y>               & optional: coordinate of the node $i \in \SV$ as a point in the left image\\
    \ttfamily i1 <s> <x> <y>               & optional: coordinate of the node $s \in \SL$ as a point in the right image\\
    \ttfamily n0 <i> <j>                   & optional: specifies that points $i, j \in \SV$ are neighbors in the left image\\
    \ttfamily n1 <s> <l>                   & optional: specifies that points $s, l \in \SL$ are neighbors in the right image\\
    \bottomrule
  \end{tabular}
  \caption{
    Short specification of the dd file format.
    The format originated in~\cite{GraphMatchingDDTorresaniEtAl}.
    It encodes a sparse representation of the graph matching problem~\eqref{equ:graph-matching-def}.
    Unary costs are encoded as \emph{assignments} and pairwise costs are encoded as \emph{edges} between two assignments.
    Any lines marked as optional or comments are ignored when building the graph matching problem.
    They were introduced in~\cite{GraphMatchingDDTorresaniEtAl} to ease visualization of matching problems between two images.
    \label{fig:dd}
  }
\end{table*}

%% file: floats/appendix-algorithm-non-positive.tex
\begin{algorithm}[t]
  \MyAlgBugFix
  \DontPrintSemicolon
  \caption{\strut Transform into non-positive costs.}
  \label{alg:appendix-non-positive}
  \KwIn{Graph matching problem $(\SV, \SL, c)$}
  \KwOut{Equivalent non-positive cost vector $c'$}

  \medskip
  \tcc{initialize}
  $c' \leftarrow (0 \ldots 0)$\;

  \medskip
  \tcc{first, shift unary costs}
  \For{$i\in\SV$ and $s\in\SL$}{
      $\alpha \leftarrow \max \{ c_{is'} \mid s' \in \SL \text{ and } c_{is'} \ne \infty \}$\;
      $c'_{is} \leftarrow c_{is} - \max \{0, \alpha\}$\;
  }
  \medskip
  \tcc{second, shift pairwise costs}
  \For{$i,j \in\SV$, $i < j$ and $s,t \in\SL$}{
      $\alpha \leftarrow \max \{ c_{is',jl'} \mid s',l' \in \SL \}$\;
      $c'_{is,jl} \leftarrow c_{is,jl} - \max\{0, \alpha\}$\;
  }
\end{algorithm}

%% file: floats/appendix-algorithm-zero-unary.tex
\begin{algorithm}[t]
  \MyAlgBugFix
  \DontPrintSemicolon
  \def\MyCaseSpace{\hspace{.4em}}%
  \caption{\strut Transform unary into pairwise costs.}
  \label{alg:zero-unary}
  \KwIn{Graph matching problem $(\SV, \SL, c)$}
  \KwOut{Equivalent cost vector $c'$ with empty diagonal}

  \medskip
  \tcc{initialize}
  \For{$i,j \in \SV$ and $s,l \in\SL$}{
    $c'_{is,jl} \leftarrow \begin{cases}
      \MyCaseSpace c_{is,jl} & \text{if } (i, s) \ne (j, l) \\
      \MyCaseSpace 0         & \text{otherwise}
    \end{cases}$\;
  }

  \medskip
  \tcc{count how many additional cost entries are necessary; remember decision with least number of additional entries in array $J$}
  \For{$i \in \SV$ and $s \in \SL$}{
    $\displaystyle
    J_{is} \leftarrow \argmin_{j \in \SV}
    |\{ l \in \SL \mid c_{is,jl} = 0 \text{ and } i \ne j, s \ne l \}|$\;
  }

  \medskip
  \tcc{distribute unaries across pairwise costs}
  \For{$i, j \in \SV$, $i < j$ and $s, l \in \SL$}{
    \If{$J_{is} = j$}{
      $c'_{is,jl} \leftarrow c'_{is,jl} + c_{is,is}$\;
    }
    \If{$J_{jl} = i$}{
      $c'_{is,jl} \leftarrow c'_{is,jl} + c_{jl,jl}$\;
    }
  }
\end{algorithm}

%% file: floats/appendix-table-cmdline.tex
\begin{table*}[t]
  \sffamily
  \let\normalfont\sffamily
  \def\DoubleDash{-\kern0pt-}%
  \def\I{\hspace*{5mm}}%
  \centering
  \begin{tabular}{l >{\vttfamily}l}
    \toprule
    \bfseries method & \normalfont\bfseries command line\\
    \midrule
    \Salg{dd-ls0}  & tkrgm \DoubleDash linear \DoubleDash tree \DoubleDash max-iter N\\
    \Salg{dd-ls3}  & tkrgm \DoubleDash linear \DoubleDash local 2 \DoubleDash tree \DoubleDash max-iter N\\
    \Salg{dd-ls4}  & tkrgm \DoubleDash linear \DoubleDash local 3 \DoubleDash tree \DoubleDash max-iter N\\
    \Salg{fw}      & graph\_matching\_frank\_wolfe\_text\_input input.dd \\
    \Salg{mp}      & graph\_matching\_mp -i input.dd \DoubleDash roundingReparametrization uniform:0.5\\
    \Salg{mp-mcf}  & graph\_matching\_mp\_tightening -i input.dd \DoubleDash tighten \DoubleDash tightenInterval 50 \textbackslash\\
                   & \I\DoubleDash tightenIteration 200 \DoubleDash tightenConstraintsPercentage 0.01 \textbackslash \\
                   & \I\DoubleDash tightenReparametrization uniform:0.5 \DoubleDash graphMatchingRounding mcf\\
    \Salg{mp-fw}   & graph\_matching\_mp\_tightening -i input.dd \DoubleDash tighten \DoubleDash tightenInterval 50 \textbackslash\\
	               & \I\DoubleDash tightenIteration 200 \DoubleDash tightenConstraintsPercentage 0.01  \textbackslash\\
				   & \I\DoubleDash tightenReparametrization uniform:0.5 \DoubleDash graphMatchingRounding fw\\
    \Salg{fm}      & qap\_dd \DoubleDash max-batches N \DoubleDash batch-size 0 \DoubleDash generate 1\\
    \Salg{fm-bca}  & qap\_dd \DoubleDash max-batches N \DoubleDash batch-size 10 \DoubleDash generate 10\\
    \bottomrule
  \end{tabular}
  \caption{
    Command line parameters for all \Cpp methods.
    \label{tab:cmdline}
  }
\end{table*}

%% file: floats/appendix-figure-performance-profile.tex
\begin{figure*}[p]
  \setlength\tabcolsep{3mm}%
  \def\MyHeight{44mm}
  \centering

  \centerline{%
  \begin{tabular}{c c}
    \includegraphics[height=\MyHeight]{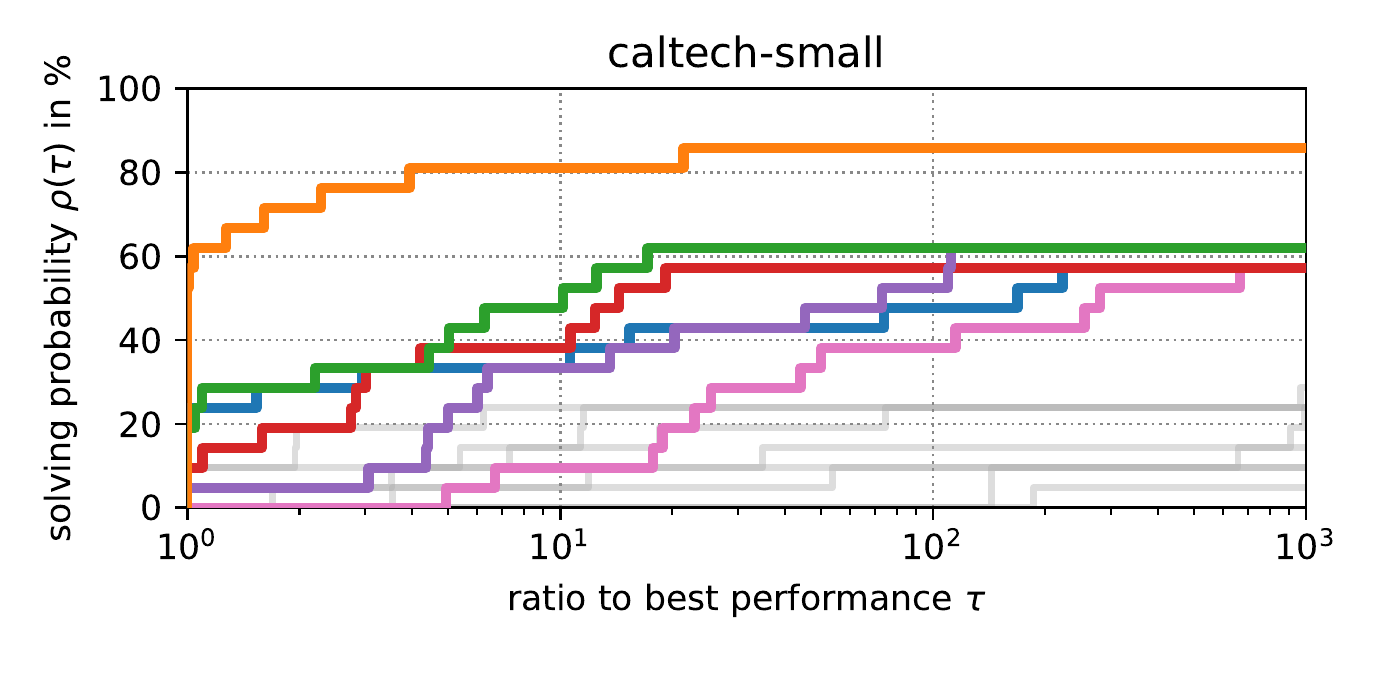} &
    \includegraphics[height=\MyHeight]{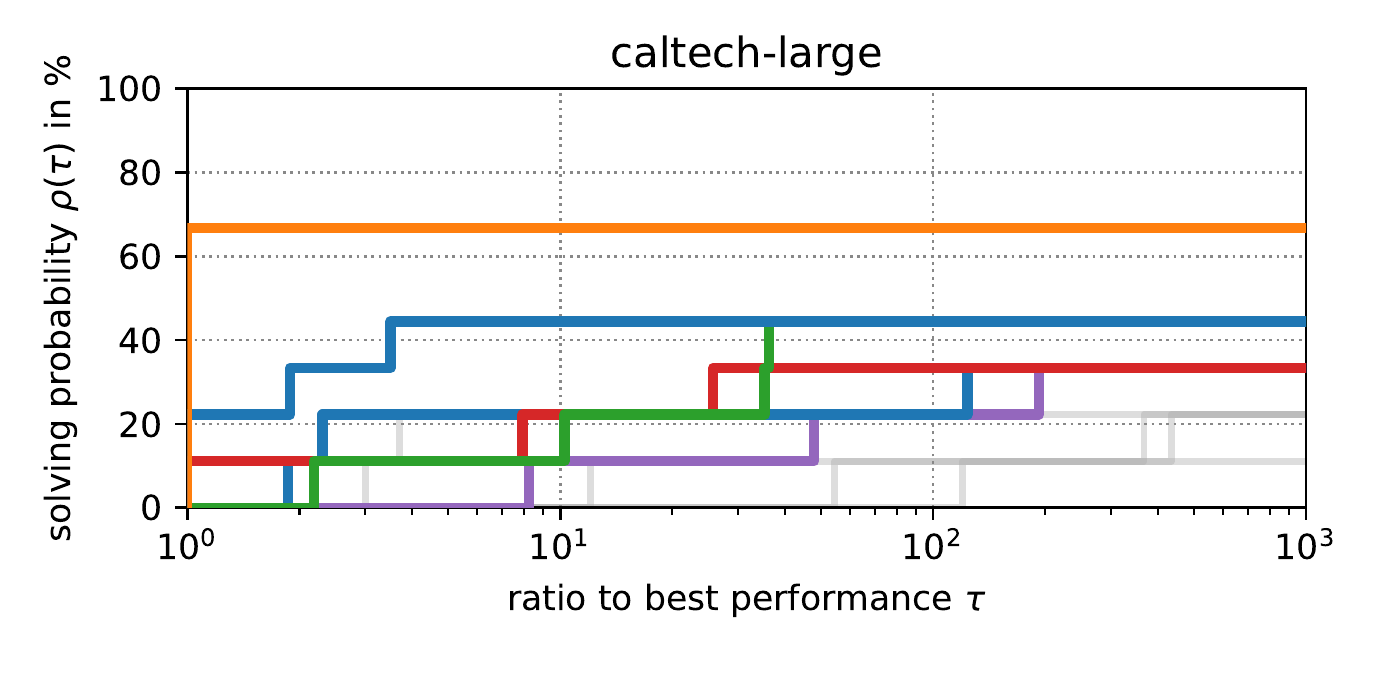} \\[-5mm]
    \includegraphics[height=\MyHeight]{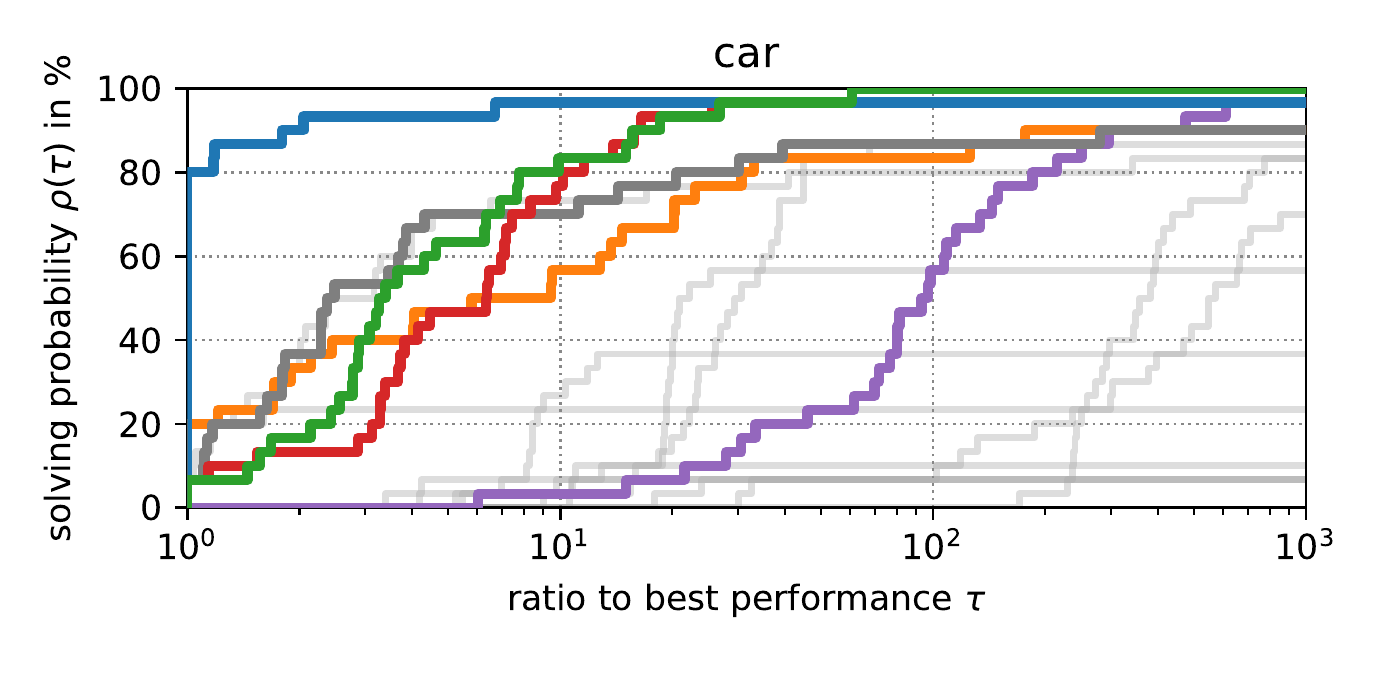} &
    \includegraphics[height=\MyHeight]{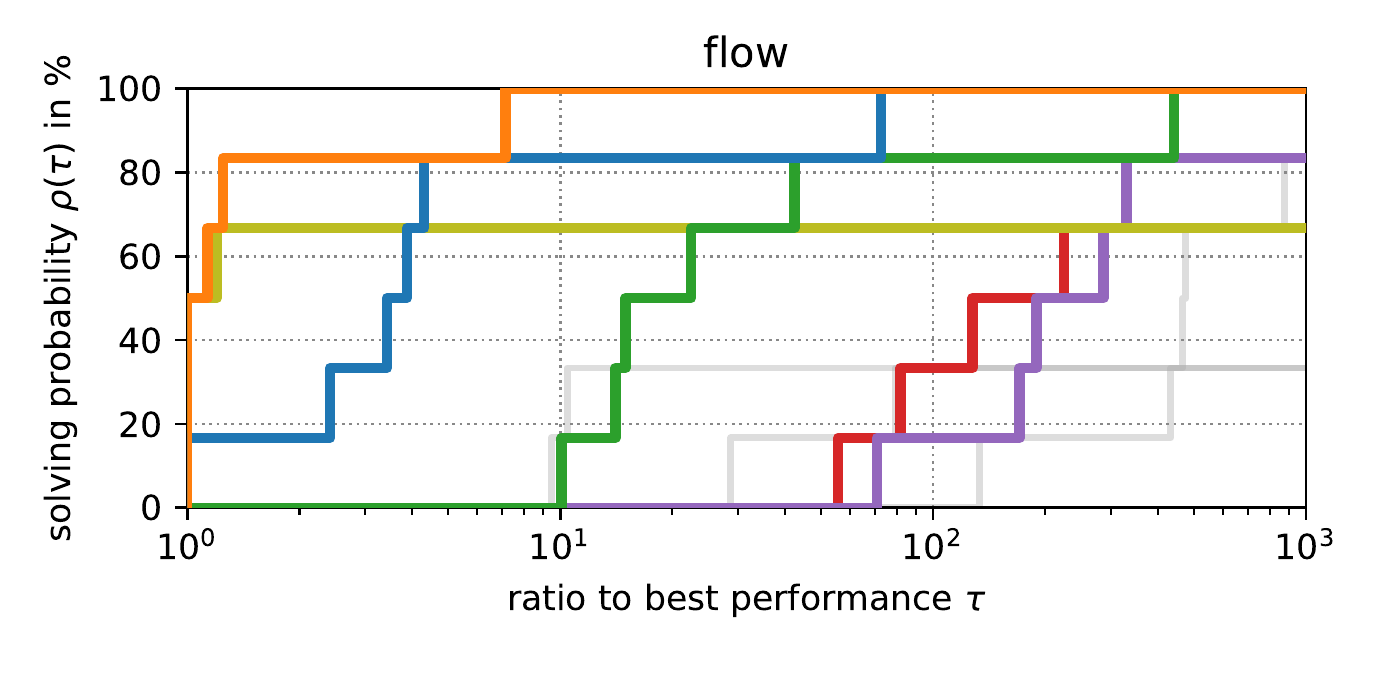} \\[-5mm]
    \includegraphics[height=\MyHeight]{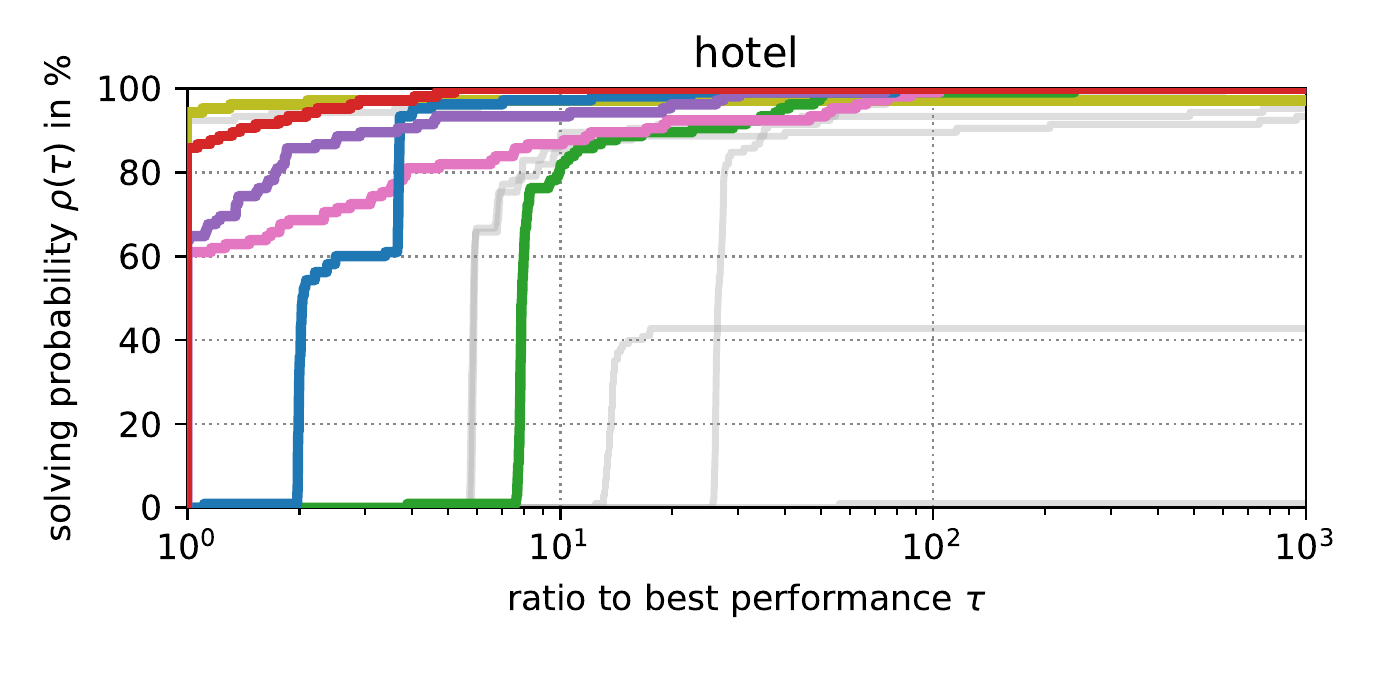} &
    \includegraphics[height=\MyHeight]{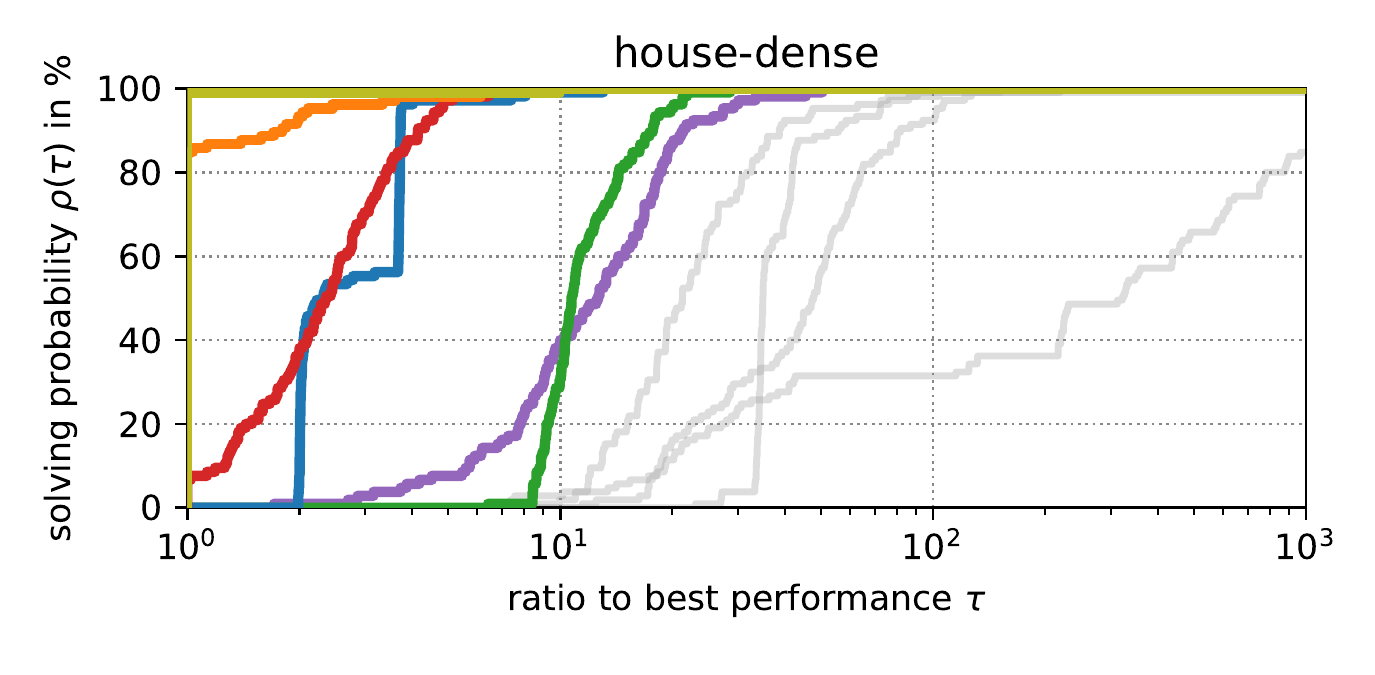} \\[-5mm]
    \includegraphics[height=\MyHeight]{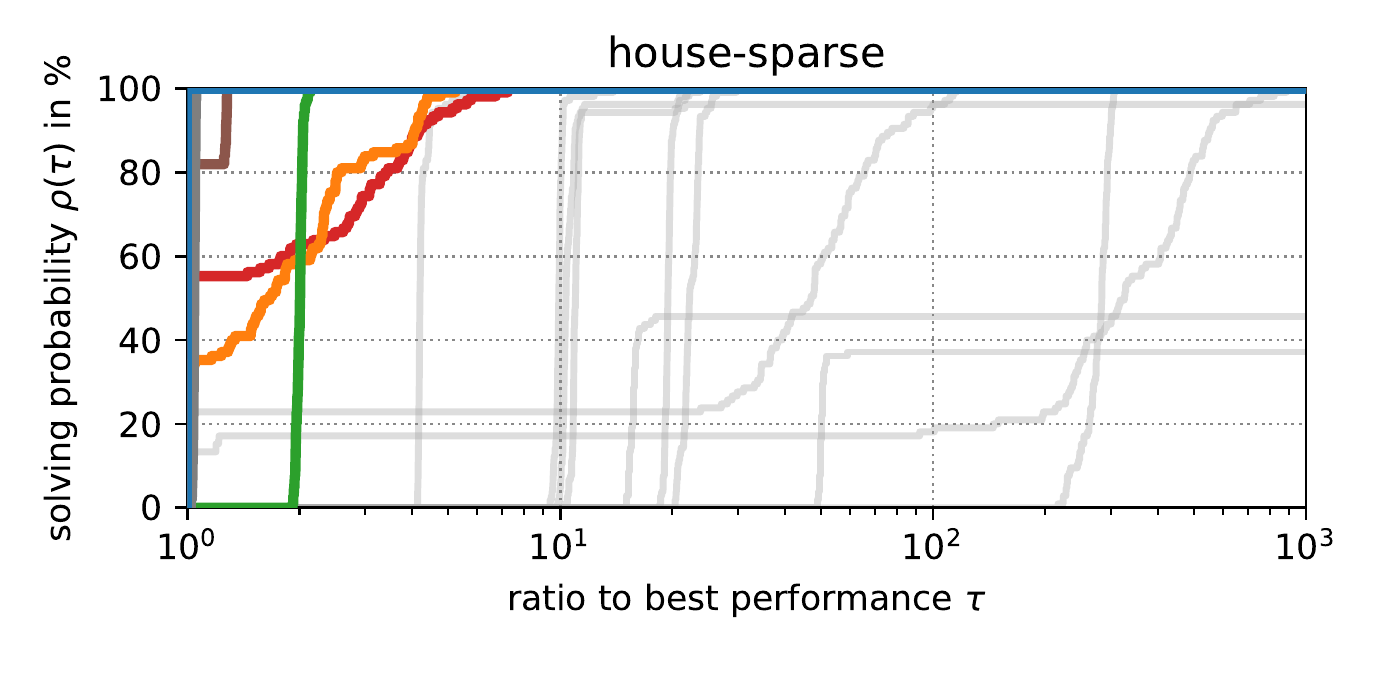} &
    \includegraphics[height=\MyHeight]{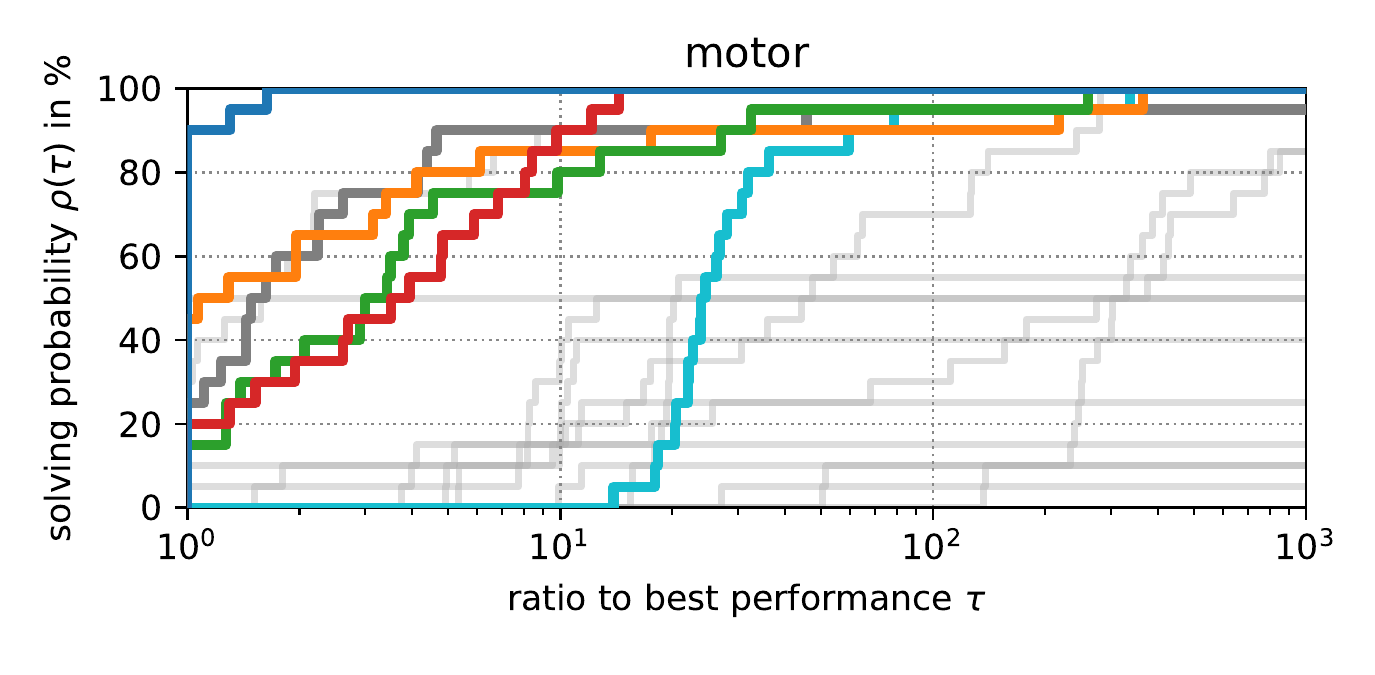} \\[-5mm]
    \includegraphics[height=\MyHeight]{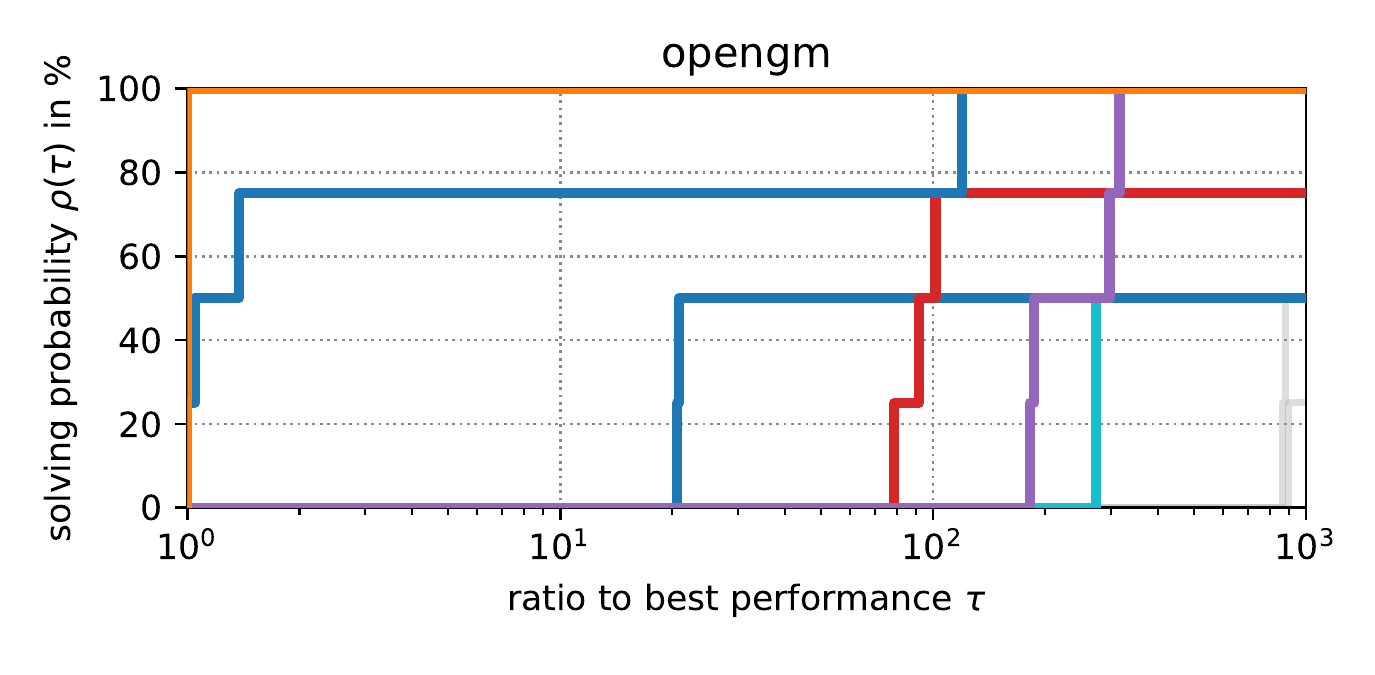} &
    \includegraphics[height=\MyHeight]{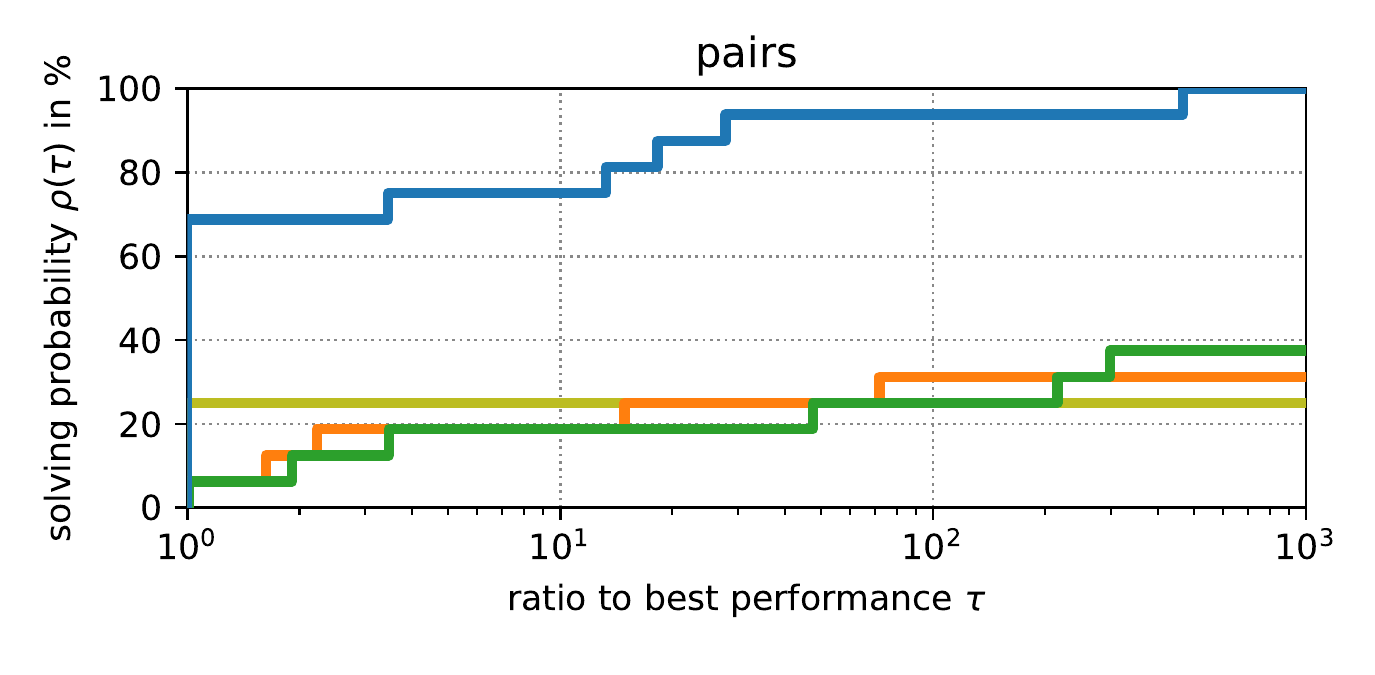} \\[-5mm]
    \includegraphics[height=\MyHeight]{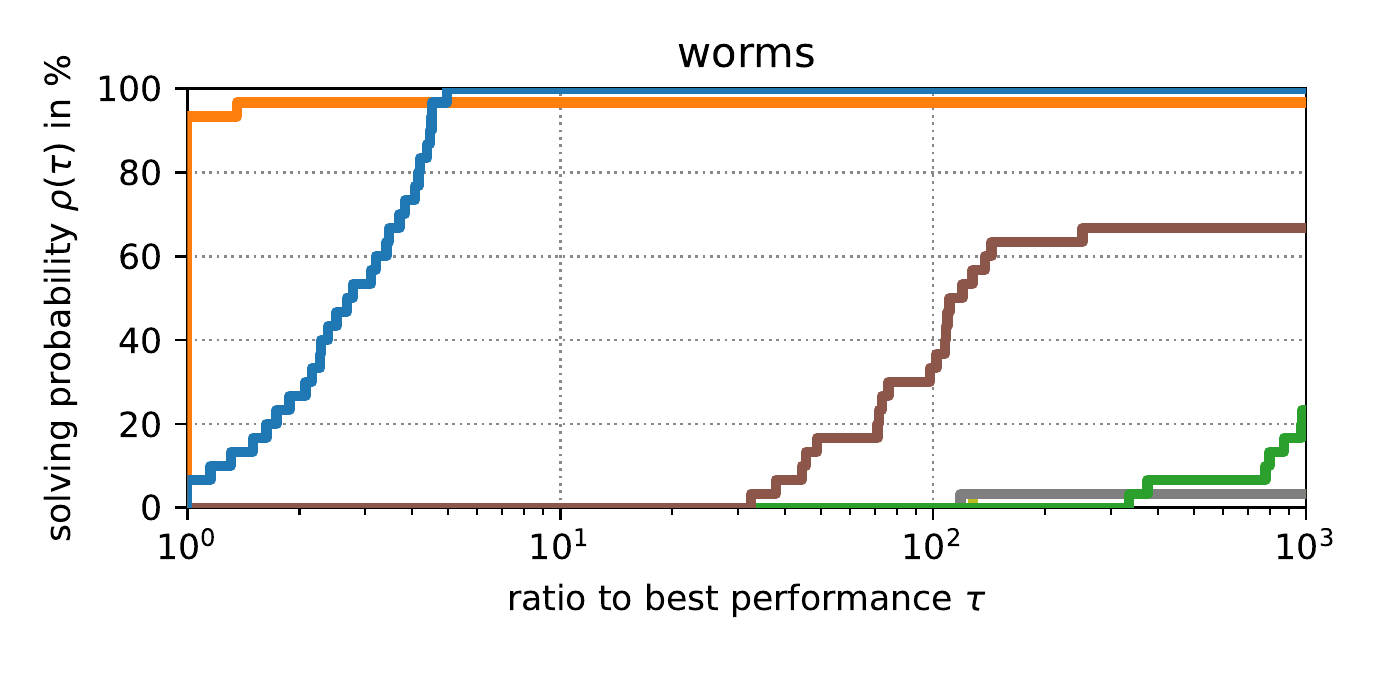} &
    \raisebox{11mm}{\hspace{7mm}\includegraphics{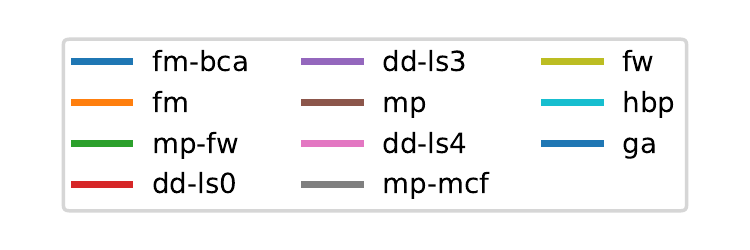}}
  \end{tabular}}
  \vspace{-5mm}

  \caption{
    Run time performance profile~\cite{dolan2002benchmarking} per dataset.
    \label{fig:appendix-performance-profile}
  }
\end{figure*}


%% file: floats/appendix-table-full-dataset.tex
\def\MyColumnNumber{\the\numexpr 2 \ifHasOpt + 1 \fi \ifHasAcc + 1 \fi}
\def\MyColumnWidth{\the\dimexpr \textwidth / (1 + 4 * \MyColumnNumber) - \tabcolsep * 2\relax}
\def\MyHead#1{\makebox[\MyColumnWidth]{#1}}
\edef\MyHeaderRow{\ifHasOpt \noexpand\MyHead{opt} & \fi \noexpand\MyHead{E} & \noexpand\MyHead{D} \ifHasAcc & \noexpand\MyHead{acc} \fi}

\begin{table*}[p]
  \smaller\sffamily
  \setlength\tabcolsep{0pt}
  \setlength\aboverulesep{0pt}
  \setlength\belowrulesep{0pt}
  \def\arraystretch{1.15}
  \centerline{%
  \begin{tabular}{l
                  *{\MyColumnNumber}{g} 
                  *{\MyColumnNumber}{c} 
                  *{\MyColumnNumber}{g} 
                  *{\MyColumnNumber}{c} 
                 }
    \toprule
    \MyHead{}
    & \multicolumn{\MyColumnNumber}{g}{\Sdata{\Dataset}\,(1s)}
    & \multicolumn{\MyColumnNumber}{c}{\Sdata{\Dataset}\,(10s)}
    & \multicolumn{\MyColumnNumber}{g}{\Sdata{\Dataset}\,(100s)}
    & \multicolumn{\MyColumnNumber}{c}{\Sdata{\Dataset}\,(300s)} \\
    & \MyHeaderRow
    & \MyHeaderRow
    & \MyHeaderRow
    & \MyHeaderRow \\
    \midrule
    \input{generated-tables/full-\Dataset} 
    \bottomrule
  \end{tabular}}

  \smallskip\raggedright
  \textbf{opt:} optimally solved instances (\%);\quad
  \textbf{E:} average best objective value;\quad
  \textbf{D:} average best lower bound if applicable;

  \textbf{acc:} average accuracy corresponding to best objective (\%);\quad
  \textbf{---*:} method yields no solution for at least one problem instance

  \caption{
    Detailed fixed-time evaluation for dataset \Sdata{\Dataset}.
    \label{tab:appendix-full-dataset-\Dataset}
  }
\end{table*}

%% file: generated-tables/full-worms.tex
\Salg{fgmd} & 0 & \multicolumn{3}{g}{---*} & 0 & \multicolumn{3}{c}{---*} & 0 & \multicolumn{3}{g}{---*} & 0 & \multicolumn{3}{c}{---*} \\
\Salg{fm} & \bfseries 93 & -48457 & -55757 & 89 & \bfseries 93 & -48458 & -55757 & 89 & \bfseries 93 & -48461 & -55757 & 89 & \bfseries 93 & -48461 & -55757 & 89 \\
\Salg{fw} & 0 & -46974 & -- & 81 & 3 & -48032 & -- & 85 & 3 & -48038 & -- & 85 & 3 & -48038 & -- & 85 \\
\Salg{ga} & 0 & \multicolumn{3}{g}{---*} & 0 & \multicolumn{3}{c}{---*} & 0 & \multicolumn{3}{g}{---*} & 0 & \multicolumn{3}{c}{---*} \\
\Salg{ipfps} & 0 & \multicolumn{3}{g}{---*} & 0 & \multicolumn{3}{c}{---*} & 0 & -1147 & -- & 1 & 0 & -1147 & -- & 1 \\
\Salg{ipfpu} & 0 & \multicolumn{3}{g}{---*} & 0 & \multicolumn{3}{c}{---*} & 0 & 0 & -- & 0 & 0 & 0 & -- & 0 \\
\Salg{lsm} & 0 & \multicolumn{3}{g}{---*} & 0 & \multicolumn{3}{c}{---*} & 0 & \multicolumn{3}{g}{---*} & 0 & \multicolumn{3}{c}{---*} \\
\Salg{mpm} & 0 & \multicolumn{3}{g}{---*} & 0 & \multicolumn{3}{c}{---*} & 0 & \multicolumn{3}{g}{---*} & 0 & \multicolumn{3}{c}{---*} \\
\Salg{pm} & 0 & \multicolumn{3}{g}{---*} & 0 & \multicolumn{3}{c}{---*} & 0 & \multicolumn{3}{g}{---*} & 0 & \multicolumn{3}{c}{---*} \\
\Salg{rrwm} & 0 & \multicolumn{3}{g}{---*} & 0 & \multicolumn{3}{c}{---*} & 0 & \multicolumn{3}{g}{---*} & 0 & \multicolumn{3}{c}{---*} \\
\Salg{sm} & 0 & \multicolumn{3}{g}{---*} & 0 & \multicolumn{3}{c}{---*} & 0 & -6453 & -- & 6 & 0 & -6453 & -- & 6 \\
\Salg{smac} & 0 & \multicolumn{3}{g}{---*} & 0 & \multicolumn{3}{c}{---*} & 0 & \multicolumn{3}{g}{---*} & 0 & \multicolumn{3}{c}{---*} \\
\midrule
\Salg{dd-ls0} & 0 & 60443 & -163870 & 26 & 0 & 50517 & -148830 & 25 & 0 & -3982 & -58449 & 58 & 0 & -43824 & -48682 & 87 \\
\Salg{dd-ls3} & 0 & 64017 & -160520 & 24 & 0 & 49257 & -144842 & 24 & 0 & 11744 & -71523 & 46 & 0 & -40882 & -48943 & 85 \\
\Salg{dd-ls4} & 0 & 65731 & -160409 & 24 & 0 & 58300 & -153566 & 25 & 0 & 31066 & -109434 & 33 & 0 & 12795 & -75088 & 44 \\
\Salg{fm-bca} & \bfseries 93 & \bfseries -48460 & \bfseries-48514 & 89 & \bfseries 93 & \bfseries -48464 & \bfseries-48498 & 89 & \bfseries 93 & \bfseries -48464 & -48498 & 89 & \bfseries 93 & \bfseries -48464 & -48498 & 89 \\
\Salg{hbp} & 0 & \multicolumn{3}{g}{---*} & 0 & \multicolumn{3}{c}{---*} & 0 & \multicolumn{3}{g}{---*} & 0 & \multicolumn{3}{c}{---*} \\
\Salg{mp} & 0 & \multicolumn{3}{g}{---*} & 67 & -48391 & \bfseries-48498 & 89 & 70 & -48393 & \bfseries-48497 & 89 & 70 & -48393 & \bfseries-48496 & 89 \\
\Salg{mp-fw} & 0 & \multicolumn{3}{g}{---*} & 3 & \multicolumn{3}{c}{---*} & 57 & -48402 & -48759 & 88 & 83 & -48435 & -48631 & 88 \\
\Salg{mp-mcf} & 0 & \multicolumn{3}{g}{---*} & 3 & -47942 & -48588 & 88 & 3 & -48027 & -48558 & 88 & 3 & -48057 & -48552 & 88 \\